\documentclass[twocolumn]{article}
\usepackage{graphicx} 
\usepackage[round, sort]{natbib}

\usepackage{enumitem}
\usepackage{hyperref} 
\usepackage{amsmath}
\usepackage{tabularx}
\usepackage{booktabs}
\usepackage{multirow}
\usepackage{appendix}
\usepackage{longtable}
\usepackage{xltabular}
\usepackage{caption}
\usepackage{csquotes}
\usepackage{makecell}
\usepackage{lipsum}
\usepackage{subcaption}
\usepackage{hhline}
\usepackage{xcolor}
\usepackage{booktabs}
\usepackage{authblk}
\usepackage{kpfonts}
\usepackage{pifont}

\begin{document}
\newcommand{\tb}[1]{\textbf{#1}} 
\newcommand{\cg}[1]{\textit{\textcolor{gray}{#1}}}
\newcommand{\cgreen}[1]{\textit{\textcolor{green}{#1}}}
\newcommand{\cred}[1]{\textbf{\textcolor{red}{#1}}}

\title{\huge\bfseries Mining the Explainability and Generalization: Fact Verification Based on Self-Instruction} 
\author[1]{\textbf{Guangyao Lu}}
\author[2]{\textbf{Yulin Liu \textsuperscript{\textasteriskcentered}}}

\affil[1,2]{College of Computer Science and Electronic Engineering, Hunan University}
\affil[1]{\href{mailto:luguangyao@hnu.edu.cn}{luguangyao$@$hnu.edu.cn}} 
\affil[2]{\href{mailto:yuling_liu@hnu.edu.cn}{yuling\_liu$@$hnu.edu.cn}} 
\date{}

\maketitle

\begin{abstract}


Fact-checking based on commercial LLMs has become mainstream. Although these methods offer high explainability, it falls short in accuracy compared to traditional fine-tuning approaches, and data security is also a significant concern. In this paper, we propose a self-instruction based fine-tuning approach for fact-checking that balances accuracy and explainability. Our method consists of Data Augmentation and Improved DPO fine-tuning. The former starts by instructing the model to generate both positive and negative explanations based on claim-expdence pairs and labels, then sampling the dataset according to our customized difficulty standards. The latter employs our proposed improved DPO to fine-tune the model using the generated samples. We fine-tune the smallest-scale Llama-7b model and evaluate it on the challenging fact-checking datasets FEVEROUS and HOVER, utilizing four fine-tuning methods and three few-shot learning methods for comparison. The experiments demonstrate that our approach not only retains accuracy comparable to, or even surpassing, traditional fine-tuning methods, but also generates fluent explanation text. Moreover, it also exhibit high generalization performance. Our method is the first to leverage self-supervised learning for fact-checking and innovatively combines contrastive learning and improved DPO in fine-tuning LLMs, as shown in the experiments.

\end{abstract}

\section{Introduction}
Fact verification, also called fact-checking, aims at determining the veracity of claims based on multiple reliable pieces of evidence\citep{glockner-etal-2022-missing,10.1162/tacl_a_00454}. It is an important task for combating the spread of misinformation on social media platforms \citep{doi:10.1126/science.aap9559,10.1145/3397271.3401396}.

Most existing fact-checking methods are based on fine-tuning pre-trained models to achieve multi-class classification models\citep{chiang-etal-2022-recent}. These methods take claims and evidence as input, which is undergone highly customized preprocessing \citep{10.1145/3539618.3592049,wang-etal-2022-qadialmoe,ma-etal-2022-glaf}, and produce probability distributions of claim veracity. Although these works achieved excellent results, the implementation methods themselves are inherently unable to meet the requirements of trustworthy fact-checking, as highlighted by \cite{pan-etal-2023-fact}: The model should not only predict the veracity of the claim but also provide a clear explanation of its reasoning process to help users understand and trust the results.

Recently, some researchers have focused their attention on large language models(LLMs)\citep{Naveed2023ACO} such as ChatGPT\citep{10.5555/3495724.3495883}, to implemented explainabe fact verification\citep{pan-etal-2023-fact,zhang-gao-2023-towards}. These methods follow the following process: Firstly, they decompose complex claims into simpler sub-claims that exhibit parallel or hierarchical relationships. Secondly, they adopt a question-and-answer approach, leveraging the intelligence of LLMs and relevant evidence, to systematically verify each sub-claim. Finally, based on the collective results of these verifications, they provide an overall assessment of the truthfulness of the complex claim.

These methods leverage the intelligence of large language models to their fullest extent, breaking the limitations of existing approaches and achieving remarkable results in few-shot and zero-shot. However, these methods also face challenges:
\begin{enumerate}[label=\textbullet, font=\footnotesize, itemsep=0pt,leftmargin=0.7em]
\item Uncontrollable cost: These methods decompose the evidence gathering task into independent or dependent sub-tasks for fine-grained reasoning, and as the granularity decreases, the size of the sub-tasks increases as well, which means a higher cost in interface invocation.
\item Overlooking the potential of open-source models: Existing methods directly utilize commercial models like ChatGPT, which raises concerns about data privacy. Moreover, there is currently limited research on transferring the inference ability of ChatGPT to smaller open-source models such as Llama-2-7b\citep{touvron2023llama}, a much smaller model with 7 billion parameters(we refer to Llama-2-7b as Llama2 for the sake of convenience), or improving the inference abilities of Llama2.
\end{enumerate}

To address these issues, we first propose a question: Can we leverage open-source LLMs, such as Llama2, to perform effective fact-checking?  Considering that existing research\citep{Hu_Sheng_Cao_Shi_Li_Wang_Qi_2024} point out that large language models excel at challenging tasks but may exhibit illusions on simpler tasks, and they act as a filter for difficult problems\citep{ma-etal-2023-large}, we believe that judging claim veracity is a simple problem, while providing relevant explanations is a difficult problem, as it requires capturing causal relationships in addition to generating explanatory content. Therefore, we attempt to provide claims, evidence and labels to the LLMs and ask it to provide reasonable explanations. Experimental results show that in most cases, the model can generate reasonable text to explain the contradictions between claims and evidence, even without adopting a decomposition approach, as demonstrated in Table \ref{tab:tip}. Furthermore, we consider jointly modeling the label of a data and the generated explanations, employing self-instruction\cite{wang2023selfinstruct} and counterfactual data augmentation\cite{zhu-etal-2023-explain} to finetune Llama2, to enhance the inference ability of the model. We also adopt PEFT\cite{peft} using an improved DPO\citep{rafailov2023direct}to fine-tune Llama2. Our contributions are as follows:
\begin{itemize}[label=\textbullet, font=\footnotesize, itemsep=0pt,leftmargin=0.7em]
\item We propose a two-dimensional data augmentation approach that involves counterfactual data augmentation and generating multiple explanations using self-instruction based method to achieve joint modeling of labels and explanations. These methods effectively expand the available data and make full use of limited data.
\item We propose a improvements method to the DPO algorithm to enhance its performance.
\item We filled the gaps in the research on self-instruction for small open-source models, introducing an efficient fine-tuning method based on DPO that enhances the inference capabilities of Llama2. Experimental results shows that our method achieves both high performance and explainability.
\end{itemize}

\section{Related Work}



\subsection{Pretrained Language Based}
In recent years, fact-checking work utilized pretrained language models to conduct a series of interesting studies focusing on retrieval, reasoning of the fact-checking workflow. For instance, \cite{sheng-etal-2021-article} proposed an improvement to information retrieval methods, taking into account the characteristics of events and patterns in fact-checking articles. \cite{wu2023enhancing} addressed structured evidence such as tables and advocated for joint training of fact verification and the selection of row and column information, thereby enhancing the model's ability to retrieve information from structured data. \cite{si-etal-2021-topic} incorporates the perception of topic and stance features and highlights the importance of topic consistency between claims and evidence, as well as the underlying stance of evidence towards the claim. \cite{Zhang_Zhang_Zhou_2024} discovered the presence of false correlations in multi-hop fact verification datasets, and proposed Casual Walk to address this problem, which introduces an intermediary inference path between the inputand the outcome, extending the inference path and compelling the model to abandon the use of false correlations as shortcuts.

Despite the excellent work mentioned above, they lack of explainability, and some efforts also focused on addressing this issue. \cite{10.1145/3292500.3330935} proposed a sentence-comment co-attention network to jointly capture interpretable top-k sentences and user comments for detecting fake news. \cite{10.1145/3539618.3592049} introduced ReRead, advocating for dual reviewing to improve the accuracy of retrieving correct evidence. Another viable approach is to adopt an ablation strategy to evaluate the salience of each reasoning path. For instance, \cite{10.1609/aaai.v37i11.26591} treated the evidence extraction task as a subgraph extraction task, where each subgraph represents a reasoning path from evidence and claims to conclusions.  Although the aforementioned methods have improved interpretability to some extent, due to the limitations of their model structures, they only provide interpretable structured information rather than generating human-readable natural language text.

\subsection{LLMs Based}
With the emergence of tools like ChatGPT\citep{ouyang2022training}, fact-checking based on LLMs has become increasingly mainstream. These methods leverage the reasoning and generation capabilities of LLMs to achieve high accuracy while generating human-readable explanatory content. Despite the illusion problem associated with LLMs, they still represent an excellent solution for achieving interpretable factual evidence.


Fact verification based on LLMs is mainly divided into two approaches: Question decomposition and Retrieval Augmented Generation \cite{lewis2020retrieval}. ProgramFC proposed by \cite{pan-etal-2023-fact} utilizes contextual learning of LLMs to decompose complex claims into multi-step reasoning programs and then performs step-by-step inference on the claims to arrive at a final conclusion. \cite{zhang-gao-2023-towards} introduces a hierarchical prompting method(HiSS) that guides LLMs to explore the sub claim in more depth. Apart from that, \cite{wang-shu-2023-explainable} proposed FOLK, which made improvements information retrieval by utilizing Google search to retrieve answers to the questions, and finally reintegrates the retrieved information back into the model to improve accuracy. \cite{chen2023complex} divides the retrieval process into two steps. Firstly, a large number of documents are extracted from Bing based on sub-questions posed by LLMs. Then, these documents are refined, summarized, and the most crucial information is extracted. \cite{Li2023SelfCheckerPM} leverages AI Agent technology \citep{yao2022react} for fact-checking, decomposing the process into tasks such as claim decomposition, information retrieval, information extraction, and evidence declaration, and treat them as planning tasks where the model fully determines the plan.


The aforementioned methods based on LLMs are excellent, they require external interfaces from OpenAI, such as text-davinci-003, leading to higher costs and compromises in terms of data security. Different from previous work, our proposed method focuses on improving the inference ability of the model using self-instruction techniques\citep{wang2023selfinstruct} and fine-tuning with small-scale open-source LLMs. This approach effectively avoids the aforementioned issues, as well as combining high accuracy with explainability.

\begin{figure*}[t]
\centering
\includegraphics[width=1\linewidth]{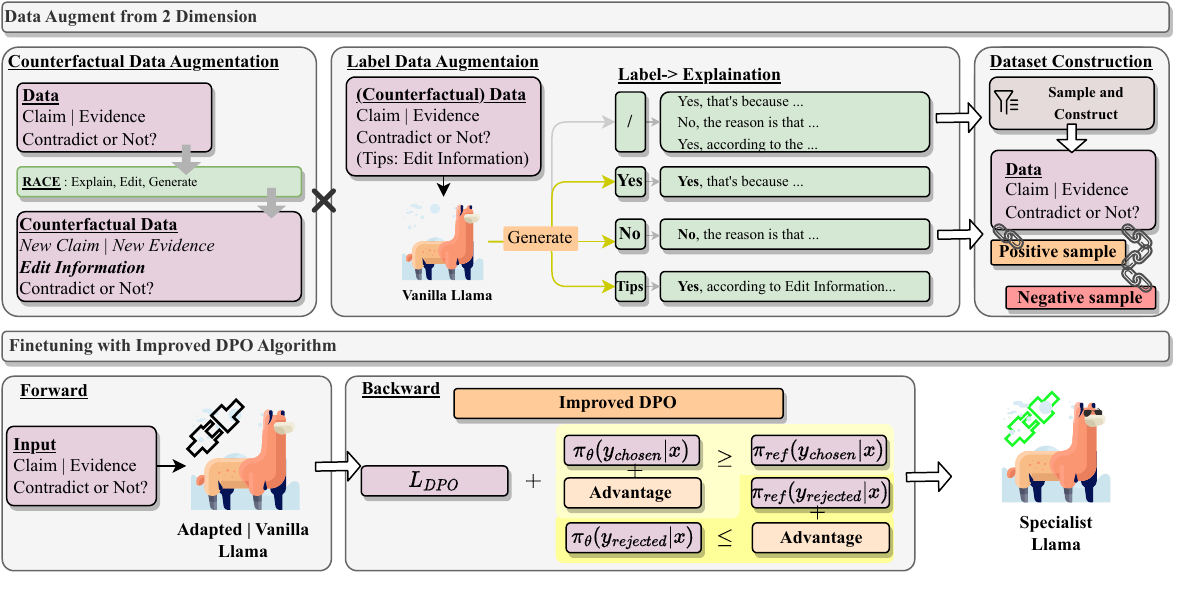}
\caption{Our approach consists of two parts: data augmentation and DPO based finetuning. The former expand the data from both the data dimension and the label dimension. The latter involves fine-tuning the model using PEFT methods, such as Lora, as well as the Improved DPO Algorithm.}
\label{fig:method}
\end{figure*}

\section{Method}


In this section, we start by describing the problem, followed by a step-by-step introduction of our proposed method, consisting of two parts: 2D Data Augmentation and Improved DPO Fine-tuning, as shown in Figure \ref{fig:method}. The former is used to generate positive and negative examples, while the improvement method proposed by the latter is used for fine-tuning the model.

\subsection{Problem Formulation}
Given a claim $c$ and evidence $k$, we construct a function $I$, to convert $c$ and $k$ to a prompt $I(c,k)$(see more in section \ref{design_prompt}). According to question, a probabilistic generative model $\pi$ aims to predict a label $y \in \{\text{Yes, No}\}$, as well as the corresponding explanation $e=\{{e_1, e_2, ..., e_{n}}\}$, where $n$ should be less than or equal to the maximum number of generated tokens:

\begin{align}
P(y, e|c,k)=\pi(y,e|I(c,k)) \label{eq:predict_y_e}
\end{align}

Unlike previous fact-checking work that based on LLMs, our designed prompt $I$ is very simple. It only provides a statement and evidence(during the data augmentation phase, it may include labels and other supplementary information to enhance model performance), as demonstrated in Table \ref{tab:tip}. The model is required to check whether there is a contradiction between the statement and evidence and provide a conclusion. Our focus is on fine-tuning the model through self-instruction, aiming to achieve commercial-level performance even with small-scale LLMs. 

\subsection{2D Data Augmentation}\label{2d}

At this stage, we expand the data volume from two dimensions: the data side and the label side. On the data side, we use counterfactual-based data augmentation methods to increase the amount of data. On the label side, we generate multiple explanations as pseudo-labels for supervised model fine-tuning.

\textbf{Counterfactual Data Augmentation}. Many advanced multi-hop fact verification methods\citep{DAS2023103219} demonstrate poor generalization performance on out-of-distribution(OOD) data. In contrast, CDA(Counterfactual Data Augmentation) achieves a generalizable dataset by increasing counterfactual instances. Specifically, this method modifies and edits entities in the text while preserving overall semantic similarity, thereby altering the causal relationships between entities in the sentences. This approach enhances the model's performance by focusing on logical relationships in the sentences, rather than relying solely on data-driven improvements.

We utilized RACE proposed by \cite{zhu-etal-2023-explain} to achieve counterfactual data augmentation. This method involves extracting causal features using an explainer, then manipulating the evidence information using an editor. Based on the causal features and manipulated evidence, a generator is used to generate counterfactual claims. The method also provides information on the locations of the manipulated information, which is highly effective for large models to identify contradictions between evidence and claims.


\begin{table*}[ht]
\scriptsize
\centering
{
\begin{tabularx}{\linewidth} { 
>{\hsize=0.2\hsize\linewidth=\hsize}X 
>{\hsize=0.2\hsize\linewidth=\hsize}X 
>{\hsize=1.6\hsize\linewidth=\hsize}X}
\hline
\multicolumn{3}{>{\hsize=\dimexpr2\hsize+2\tabcolsep+\arrayrulewidth\relax}X}{
\textbf{Claim}:\newline During Washington's first district primary for the U.S. House of Representatives in 2010, Democratic candidate Jay Inslee earned 55.85 percent of the vote.
}\\\hline

\multicolumn{3}{>{\hsize=\dimexpr2\hsize+2\tabcolsep+\arrayrulewidth\relax}X}{
\textbf{Evidence}:\newline 
[Jay Inslee] Jay Robert Inslee(born February 9, 1951) is an American politician who has served as the 23rd since 2013.\newline
[Jay Inslee] In the general election, Inslee defeated Carmichael 52\%-48\%.\newline
[Jay Inslee] In 2010 he won by a 15-expoint margin, with 57.67\% of the votes cast in his favor.
}\\\hline
\multicolumn{3}{>{\hsize=\dimexpr2\hsize+2\tabcolsep+\arrayrulewidth\relax}X}{
\underline{\textbf{given - , predict $y$, $e$}}:\newline
No, the evidence does not contradict the claim. In fact, the evidence supports the claim. The claim …\newline
No, the evidence does not contradict the claim. In fact, it supports the claim. The evidence provided …\newline
Yes, I can help you with that! The evidence you provided does contradict the claim made in the …\newline
Yes, I can help you with that. The evidence you provided does indeed contradict the claim…
}
\\\hline
\multicolumn{3}{>{\hsize=\dimexpr2\hsize+2\tabcolsep+\arrayrulewidth\relax}X}{
\underline{\textbf{given $y = No$, predict $e$}}:\newline
No, the evidence does not contradict the claim. In fact, the evidence supports the claim. The claim states that during Washington's first district primary for the U.S. House of Representatives in 2010…, the evidence provided supports the claim that Jay Inslee earned 55.85\% of the vote during Washington's first district primary in 2010.
}
\\\hline
\multicolumn{3}{>{\hsize=\dimexpr2\hsize+2\tabcolsep+\arrayrulewidth\relax}X}{
\underline{\textbf{given $y = Yes$, predict $e$}}:\newline
Yes, I can help you with that. The evidence does contradict the claim in two ways:1. Year difference: The evidence states that Inslee won the general election in 2010, but the claim is about the primary election in 2010. …, \textbf{the evidence provided contradicts the claim that Inslee earned 55.85\% of the vote in the primary election in 2010.}
}
\\\hline
\end{tabularx}
}
\caption{An example showcasing the generated results of Llama2. Without providing $y$, the model lacks confidence in its predictions. When the correct $y$ is given, the model is capable of generating more reasonable explanations, as indicated by the bolded text. Conversely, given incorrect $y$ the model may also produce fluent but unreasonable explanations.}
\label{tab:tip}
\end{table*}


\textbf{Label Augmentation}. In addition to data augmentation from the data side, we propose a method to augment data from the label side. Our idea stems from observations of LLMs, as shown in Table \ref{tab:tip}. The model does not perform well in simple classification tasks, as shown by the varying labels it produces when given the same claim and evidence. On the contrary, the model may excel at complex reasoning tasks, as it has a higher likelihood of generating correct logic when given the correct label. We speculate that this is related to the model's pretrained strategy, and a similar phenomenon has been observed in the task of event extraction\citep{ma-etal-2023-large}. Therefore, we consider whether we can use the logic generated by the model as pseudo-labels for joint training with the classification labels. This approach ensures both the accuracy of the model's predictions and its explainability. Hence, we propose the following method to enhance the label side, constructing a joint representation of classification labels and explanations leveraging the reasoning capability of the model itself.

We employ the following four configurations to enhance the label side, preparing for subsequent training tasks:

\begin{align}
P(e_{yes}|c,k,y_{yes})=\pi(e|I(c,k,y_{yes}))\label{eq:given_yes_predict_e}
\end{align}
\begin{align}
P(e_{false}|c,k,y_{no}=\pi(e_{false}|I(c,k,y_{no}))\label{eq:given_no_predict_e}
\end{align}
\begin{align}
P(e_{tips}|c,k,y,tips)=\pi(e_{tips}|I(c,k,tips,y))\label{eq:given_tips_predict_y_e}
\end{align}

The first setting involves only inputting a claim $c$ and evidence $k$ and requires the model to generate a label $y$ and an explanation $e$, as shown in equation \eqref{eq:predict_y_e}, which is used to evaluate the difficulty of data points. The second setting assumes there existing contradiction between the claim and the evidence. We assign the label $y_{yes}$ and require the model to output the corresponding explanation $e_{yes}$, as shown in Equation \eqref{eq:given_yes_predict_e}, while the third setting is the opposite, as shown in Equation \eqref{eq:given_no_predict_e}. The fourth setting takes effect only when the data is sourced from a counterfactual subset and there is a contradiction between the claim and the evidence. We provide the editing cues from the counterfactual data as tips to the model, as shown in Equation \eqref{eq:given_tips_predict_y_e}. These editing cues indicate which parts of the claim/evidence have been modified, resulting in a contradiction. We require the model to identify the points of contradiction from these editing cues in order to provide more accurate explanations.

\label{sec:training_data_construction}

\textbf{Training Data Construction}. At this stage, we construct dataset for DPO training based on the aforementioned work. Our idea is inspired by the high similarity between DPO and contrastive learning(CL)\citep{zhang-etal-2022-contrastive-data}, both of which learn through the distributional features of positive and negative samples. However, the difference is that in CL, each training data sample contains multiple positive and negative examples to help the model better capture the data distribution, while the current DPO related dataset\citep{bai2022training} only contain one positive and one negative example per sample, which cannot help the model better grasp the data distribution. Therefore, we propose a new strategy of difficulty-based sampling to generate training data, which is outlined below.


For each piece of data, we first generate $k$ different results based on setting 1, count the number of correct predictions, denoted as $w$, which helps assess the difficulty of the data. Next, We determine the number of samples $N_{sample}$ base on the equation \eqref{eq:sample}, where $N_{min}$ ensures the minimum number of samples, and $N_{base}$ is the based sampling number. Then, we randomly sample one result from both the results generated from Setting 2 and Setting 3. If the sample comes from the counterfactual subset with a label of \enquote{refutes}, we sample a correct result from Setting 4. We concatenate the prompt, label, and result to generate a positive sample $y_{chosen}$, and a negative sample $y_{rejected}$. This step is repeated for a total of $N_{sample}$ times. Finally, all the positive and negative sample pairs are saved to the training dataset for DPO training.

\begin{align}
N_{sample}=\max\Big({N_{min}, \big\lceil N_{base}*\frac{w}{k}}\big\rceil\Big) \label{eq:sample}
\end{align}

\subsection{Improved DPO Fine-tuning}
\label{dpo_fine-tuning}
In this stage, we use DPO to optimize LLMs, and the original loss function of DPO is shown in equation \eqref{eq:DPO}, where $\pi_{ref}$ refers to the basic model, $\beta$ is a hyperparameter, and $\pi_{\theta}$ represents the model enhanced with the PEFT method with trainable parameters $\theta$ injected into each layer of the transformer architecture of the model.

\begin{align}
L_{DPO}(\pi_{\theta};\pi_{ref}) =&-\Big [
\log \sigma \Big(\beta \log\frac{\pi_{\theta}(y_{chosen}|x)}{\pi_{ref}(y_{chosen}|x)}
\notag \\
&-\beta \log\frac{\pi_{\theta}(y_{rejected}|x)}{\pi_{ref}(y_{rejected}|x)} \Big)
\Big ] \label{eq:DPO}
\end{align}

It is worth noting that we observed a side effects of DPO in our experiments(see details in Section \ref{sec:poid}). To address this issue, we introduced two constraint terms $C_{chosen}$ and $C_{rejected}$ to enhance the loss function, and used two advantage values $a_1$ and $a_2$ to respectively emphasize how much $\mathcal{P}(y_{chosen})$ has increased and $\mathcal{P}(y_{rejected})$ has decreased compared to the original, as shown in equation \ref{eq:constrate}.

\begin{align}
C_{chosen}=-\log\frac{\pi_{\theta}(y_{chosen}|x)}{\pi_{ref}(y_{chosen}|x)}+log(1+a_1) \notag \\
C_{rejected}=\log\frac{\pi_{\theta}(y_{rejected}|x)}{\pi_{ref}(y_{rejected}|x)}-log(1-a_2)\label{eq:constrate}
\end{align}

After introducing the two constraint terms, we formulated it as a constrained optimization problem with inequality constraints:

\begin{align}
&\mathop{argmin}\limits_{\theta} \space L_{DPO} \notag \\
s.t. &
C_{chosen}\leq 0 \notag \\
&C_{rejected} \leq 0
\end{align}

We employ the Lagrange Multiplier\citep{bertsekas2014constrained} to convert the aforementioned constrained problem into an unconstrained problem, allowing us to directly minimize the objective function, as shown in Equation \eqref{eq:dpo_constrate}, where $\mu_1$ and $\mu_2$ are the learnable lagrange multipliers corresponding to $C_{chosen}$ and $C_{rejected}$, respectively. We multiply the constraint terms by $\beta$ and take the logarithm of their sigmoid values to ensure that the scale of the constraint terms is consistent with that of $L_{DPO}$:

\begin{align}
    Loss=&L_{DPO} + \log \sigma \Big[ \beta(\mu_1 C_{chosen} +\mu_2 C_{rejected})\Big] \label{eq:dpo_constrate}
\end{align}

To ensure that the aforementioned optimization problem satisfies the Karush-Kuhn-Tucker(KKT) conditions, we adopt the following approach since deep learning networks are not convex functions. For $\theta$, we update it using gradient descent. As for the lagrange multipliers, we update them as equation \eqref{eq:lagrange_mutilplier}, where $lr_\mu$ represents the learning rate for the Lagrange multipliers. We adhere to the KKT conditions, which require the constraints $u_1$ and $u_2$ to be greater than or equal to 0.

\begin{align}
\mu_1=max(\mu_1 - lr_{\mu}*C_{chosen} , 0) \notag \\
\mu_2=max(\mu_2 - lr_{\mu}*C_{rejected} , 0) \label{eq:lagrange_mutilplier}
\end{align}

\section{Experiments}
\subsection{Datasets}

We evaluate the performance of our model using two datasets, FEVEROUS and HOVER. 

\textbf{FEVEROUS} is proposed by \cite{aly-etal-2021-fact}, consisting of fact-checking data with various challenges such as Numerical Reasoning, Entity Disambiguation, and Multi-hop Reasoning.


\textbf{HOVER} was proposed by \cite{yang-etal-2018-hotpotqa}, which serves as a more challenging multi-hop fact-checking dataset, with a maximum hop count of 4. This is attributed not only to the increased number of hops but also to the longer evidence provided, most of which contain distracting information. Appendix \ref{section:sod} presents detailed statistical data for the two datasets.

\subsection{Baselines}
\newcommand{\Rone}{\textbf{\uppercase\expandafter{\romannumeral1}}}
\newcommand{\Rtwo}{\textbf{\uppercase\expandafter{\romannumeral2}}}
\newcommand{\Rthree}{\textbf{\uppercase\expandafter{\romannumeral3}}}

We employed the following two categories of methods as our baselines:


\Rone.\space\textbf{Supervised Fine-tuning}: These methods provide binary classification results based solely on the input data to determine if there is a contradiction between the claim and the evidence. We fine-tuned these methods using the full dataset, including counterfactual data, to evaluate the advantages of our approach compared to traditional methods. We selected the following four methods: Bert-FC\citep{10.1007/978-3-030-45442-5_45} and T5\citep{jiang-etal-2021-exploring-listwise} employ pre-trained models of different sizes for binary fact-checking classification tasks.
RoBERTa-NLI\citep{nie-etal-2020-adversarial} and DeBERTa-NLI\citep{sileo2023tasksource} are fine-tuned on other textual entailment tasks. For detailed information, please refer to the appendix.

\Rtwo.\space\textbf{Few-shot Learning}: These methods leverage LLMs to generate prediction results along with relevant explanations. This includes commercially available LLMs that are difficult to fine-tune due to their large parameter volume. To tackle this, we adopt a few-shot learning approach by selecting 10 examples from the training data and utilizing their labels and explanations as prompts. This category of methods is employed to assess the performance and interpretability advantages of our model in comparison to large-scale LLMs. We have chosen the following methods for evaluation: FLAN-T5\citep{chung2024scaling}, ChatGPT-3.5\citep{ouyang2022training}, and ProgramFC\citep{pan-etal-2023-fact}, which translates the reasoning process into a program.

\subsection{Experimental Settings}

In the data processing stage, we set the maximum token count to 512 to ensure diversity. For the 4 settings mentioned above, we attempted to generate 10 different results for each. During the filtering process, we evaluated the win\_rate of the data under setting 1 and used it to determine the sampling probability. This resulted in sampling 148,232 and 221,373 preference data pairs on FEVEROUS and HOVER, respectively.

In the model fine-tuning stage, we used implemented improve DPO based on trl\citep{vonwerra2022trl}. We fine-tuned the half-precision model with 2 Nvidia RTX 4090 GPUs, setting the learning rate to 1e-4, warm-up to 200, learning decay rate to 0.999, $\beta$ to 0.1, $a_1$ to 0.1 and $a_2$ to 0.1. We used a batch size of 20 and trained for 30,000 iterations, employing an early stopping strategy. DeepSpeed was also used to manage memory consumption.

\section{Results and Discussion}
\label{sec:result_and_discussion}

\newcommand{\pfc}{\citep{pan-etal-2023-fact}}
\newcommand{\bfc}{\citep{10.1007/978-3-030-45442-5_45}}
\newcommand{\lts}{\cite{jiang-etal-2021-exploring-listwise}}
\newcommand{\rob}{\cite{nie-etal-2020-adversarial}}
\newcommand{\deb}{\cite{sileo2023tasksource}}
\newcommand{\gpt}{\cite{ouyang2022training}}
\newcommand{\flt}{\cite{chung2024scaling}}
\newcommand{\lla}{\cite{touvron2023llama}}
\newcommand{\FIR}{\textbf{\textsuperscript{1}}}
\newcommand{\SEC}{\textbf{\textsuperscript{2}}}
\newcommand{\THI}{\textbf{\textsuperscript{3}}}

\begin{table*}[t]
{
\scriptsize
\begin{tabularx}{\linewidth}{
>{\hsize=0.01\hsize\linewidth=\hsize}X 
>{\hsize=0.22\hsize\linewidth=\hsize}X| 
>{\hsize=0.1\hsize\linewidth=\hsize}X 
>{\hsize=0.1\hsize\linewidth=\hsize}X
>{\hsize=0.1\hsize\linewidth=\hsize}X 
>{\hsize=0.1\hsize\linewidth=\hsize}X 
>{\hsize=0.1\hsize\linewidth=\hsize}X 
>{\hsize=0.15\hsize\linewidth=\hsize}X| 
>{\hsize=0.1\hsize\linewidth=\hsize}X 
>{\hsize=0.1\hsize\linewidth=\hsize}X
>{\hsize=0.1\hsize\linewidth=\hsize}X 
>{\hsize=0.1\hsize\linewidth=\hsize}X 
>{\hsize=0.1\hsize\linewidth=\hsize}X 
>{\hsize=0.1\hsize\linewidth=\hsize}X 
>{\hsize=0.1\hsize\linewidth=\hsize}X 
>{\hsize=0.1\hsize\linewidth=\hsize}X}
\toprule

\multicolumn{2}{c|}{\multirow{2}{*}{\tb{METHOD}}} &\multicolumn{2}{c}{\tb{NR}}   &\multicolumn{2}{c}{\tb{ED}}   &\multicolumn{2}{c|}{\tb{FEVEROUS}}  &\multicolumn{2}{c}{\tb{2\_HOP}}   &\multicolumn{2}{c}{\tb{3\_HOP}}  &\multicolumn{2}{c}{\tb{4\_HOP}} &\multicolumn{2}{c}{\tb{HOVER}}\\ 
\multicolumn{2}{l|}{}                         &\tb{ACC}&\tb{F1} &\tb{ACC}&\tb{F1} &\tb{ACC}&\tb{F1} &\tb{ACC}&\tb{F1} &\tb{ACC}&\tb{F1} &\tb{ACC}&\tb{F1} &\tb{ACC}&\tb{F1}\\ \midrule
\multirow{4}{*}{\Rone}             & BERT-FC  &\cg{84.47}&\cg{78.87}&\cg{80.36}&\cg{80.13}&\cg{90.01}&\cg{90.00}&\cg{78.86}&\cg{78.74}&\cg{80.60}&\cg{80.25}&\cg{79.21}&\cg{79.18}&\cg{79.75}&\cg{79.57}\\
                                   & LisT5    &\cg{93.20}&\cg{88.93}&\cg{84.82}&\cg{84.81}&\cg{93.89}&\cg{93.87}&\cg{81.06}&\cg{79.31}&\cg{84.35}&\cg{84.35}&\cg{79.98}&\cg{79.98}&\cg{82.24}&\cg{82.12}\\ 
                                   & Ro-NLI   &\cg{90.29}&\cg{83.87}&\cg{77.68}&\cg{77.59}&\cg{91.63}&\cg{91.57}&\cg{84.53}&\cg{84.48}&\cg{84.85}&\cg{84.74}&\cg{83.54}&\cg{83.54}&\cg{84.42}&\cg{84.41}\\
                                   & De-NLI   &\cg{90.29}&\cg{86.01}&\cg{83.04}&\cg{82.97}&\cg{94.00}&\cg{94.02}&\cg{80.20}&\cg{80.10}&\cg{81.63}&\cg{81.29}&\cg{78.92}&\cg{78.88}&\cg{80.53}&\cg{80.35}\\ \midrule
\multirow{4}{*}{\Rtwo}             & FLAN-T5  & 38.83    & 38.83    & 54.46    & 46.59    & 61.10    & 56.21    & 41.65    & 31.52    & 49.86    & 39.11    & 49.75    & 35.90    & 47.52    & 36.17    \\
                                   & Llama2   & 44.66    & 44.47    & 60.71    & 58.82    & 65.50    & 63.64    & 53.82    & 52.58    & 54.66    & 54.63    & 50.72    & 48.69    & 53.40    & 53.39    \\ 
                                   & ChatGPT  & 68.93    & 55.21    & 74.10    & 71.35    & 73.57    & 73.36    & 55.68    & 55.61    & 52.21    & 52.21    & 54.57    & 54.16    & 53.80    & 53.90    \\
                                   & ProgramFC&91.50\SEC & 87.58\SEC&85.71\SEC & 85.42    & 91.66    & 91.59    & 76.38    & 76.33    & 69.97    & 69.26    & 66.70    & 66.60    & 70.93    & 70.65    \\\midrule 
\multirow{7}{*}{\Rthree}           & basic    &\tb{92.23\FIR}&\tb{88.80\FIR}&85.71\SEC &85.70\SEC &\bf{94.43\FIR}&\bf{94.42\FIR} &\bf{82.42\FIR}&\bf{82.41\FIR}&\bf{81.91\FIR}&\bf{81.76\FIR}&78.54\SEC&\bf{78.51\FIR}&\bf{81.30}\FIR&\bf{81.26\FIR}\\ \cmidrule(lr){2-16}
                                   & -exp     &\cg{90.29}&\cg{85.55}&\cg{86.61}&\cg{86.58}&\cg{94.46}&\cg{94.46}       &\cg{82.77}&\cg{82.73}&\cg{83.14}&\cg{83.01}&\cg{79.89}&\cg{79.88}&\cg{82.65}&\cg{82.59}\\
                                   & -cou     & - & - & - & - & - & -                                                  &81.97\THI &81.90\THI &80.27\THI &79.97\THI &77.58\THI &77.57\THI &80.05\SEC &79.93\THI \\ 
                                   & -sam     &91.26\THI & 87.02\THI&\bf{86.61\FIR} &\bf{86.60\FIR}& 94.39\SEC&94.39\SEC& 76.02    & 75.61    & 78.47    & 78.13    & 73.53    & 73.47    & 76.50    & 76.34    \\ \cmidrule(lr){2-16}
                                   & -adv     & 90.29    & 86.42    & 84.82    & 84.76    & 93.89    & 93.89            & 79.48    & 79.47    & 79.02    & 78.90    & 73.63    & 73.54    & 77.75    & 77.73    \\ 
                                   & -lag     & 90.29    & 85.55    & 85.7     & 85.70    & 93.96    & 93.95            &82.06\SEC & 82.06\SEC & 80.38\SEC & 80.24\SEC&\bf{79.02\FIR}&78.15\SEC &78.15\THI &80.25\SEC \\
                                   & -mug     & 91.26    &87.02\THI & 83.93    & 83.91    &94.23\THI & 94.22\THI        &78.15     & 78.04    & 76.40    & 76.26    & 70.93    & 70.75    & 75.47    & 75.44    \\
                                   & -dpo     & 87.38    & 82.59    & 84.82    & 84.82\THI& 92.20    & 92.19            &61.19     & 60.52    & 57.82    & 56.84    & 54.96    & 54.94    & 58.02    & 57.49    \\ \bottomrule
                                   
\end{tabularx}
\caption{The performance of baseline(\Rone, \Rtwo) and our method(\Rthree) on FEVEROUS and HOVER. Two specific challenges on Feverous: Numerical Reasoning(NR) and Entity Disambiguation(ED) and different hops on HOVER are reported. The \cg{gray font} indicates the corresponding method is a binary classification model that lacks explainability. \Rone: supervided fine-tuning methods. \Rtwo: few-shot learning methods. \Rthree: ablation experiments refering to Section \ref{sec:ablation_study}. For explainable methods, we annotate the ranking of the top three with superscripts and bold the best result.}
\label{tab:main_result}
}
\end{table*}
In this section, we present and analyze the performance of the model on two datasets. To provide a comprehensive understanding of our findings, we organize the content into four subsections. Within each subsection, we begin by introducing relevant questions and subsequently delve into a discussion and examination of these questions, drawing upon our experimental results to provide answers.


\subsection{Main Results}
Based on the experimental results presented in Table \ref{tab:main_result}, we analyzed the advantages and disadvantages of our method in comparison to traditional SFT methods and LLMs based methods. We have summarized the following two key points.


\textbf{We achieved comparable or even superior performance to traditional SFT methods by using fewer training parameters.} We provide a detailed report of the parameter distribution of the fine-tuned model in Table \ref{tab:parameter}. It can be observed that our method has an order of magnitude fewer trainable parameters compared to traditional SFT methods(only 61.5\% of Bert-FC and 18.9\% of Ro-NLI). Our method outperforms the approach used in \Rone on FEVEROUS. However, on HOVER, our performance is generally not as good as RoBERTa-NLI. 


We analyzed the experiments and identified three reasons for this: 1)Traditional SFT methods may take shortcuts by learning biases in the dataset, such as annotations, for complex multi-hop reasoning tasks, rather than learning the logical reasoning itself \citep{Zhang_Zhang_Zhou_2024}. Moreover, the length of the 4\_hops data in HOVER even exceeds the maximum processing length of \Rone, which makes the model more biased towards the data. The above analysis partly explains why \Rone\space achieves good results in the Entity Disambiguation task and the HOVER dataset. 2)Our proposed training scheme is more challenging as it requires the model to simultaneously consider labels and explaination, so compromising on accuracy to some extent is acceptable.

\textbf{Our method outperforms approaches based on LLMs and prompt engineering techniques.} Compared to ChatGPT-3.5, our method achieves f1 score improvements of 21.06\% and 27.36\% on FEVEROUS and HOVER, respectively, with significant improvements in each subtask. Compared to the carefully designed ProgramFC, our method also achieves an overall improvement of 1.73\% and 9.90\%, respectively, and its performance on  HOVER is more stable, with no drastic decline in experimental performance as the number of hops increases. This comparative experiment confirms that our proposed method can significantly enhance the performance of open-source models like Llama2, enabling them to surpass the performance of commercial large-scale models in fact-checking tasks.

\subsection{Ablation Study}
\label{sec:ablation_study}

\begin{figure}[t]
\begin{subfigure}{0.48\linewidth}
\includegraphics[width=\linewidth]{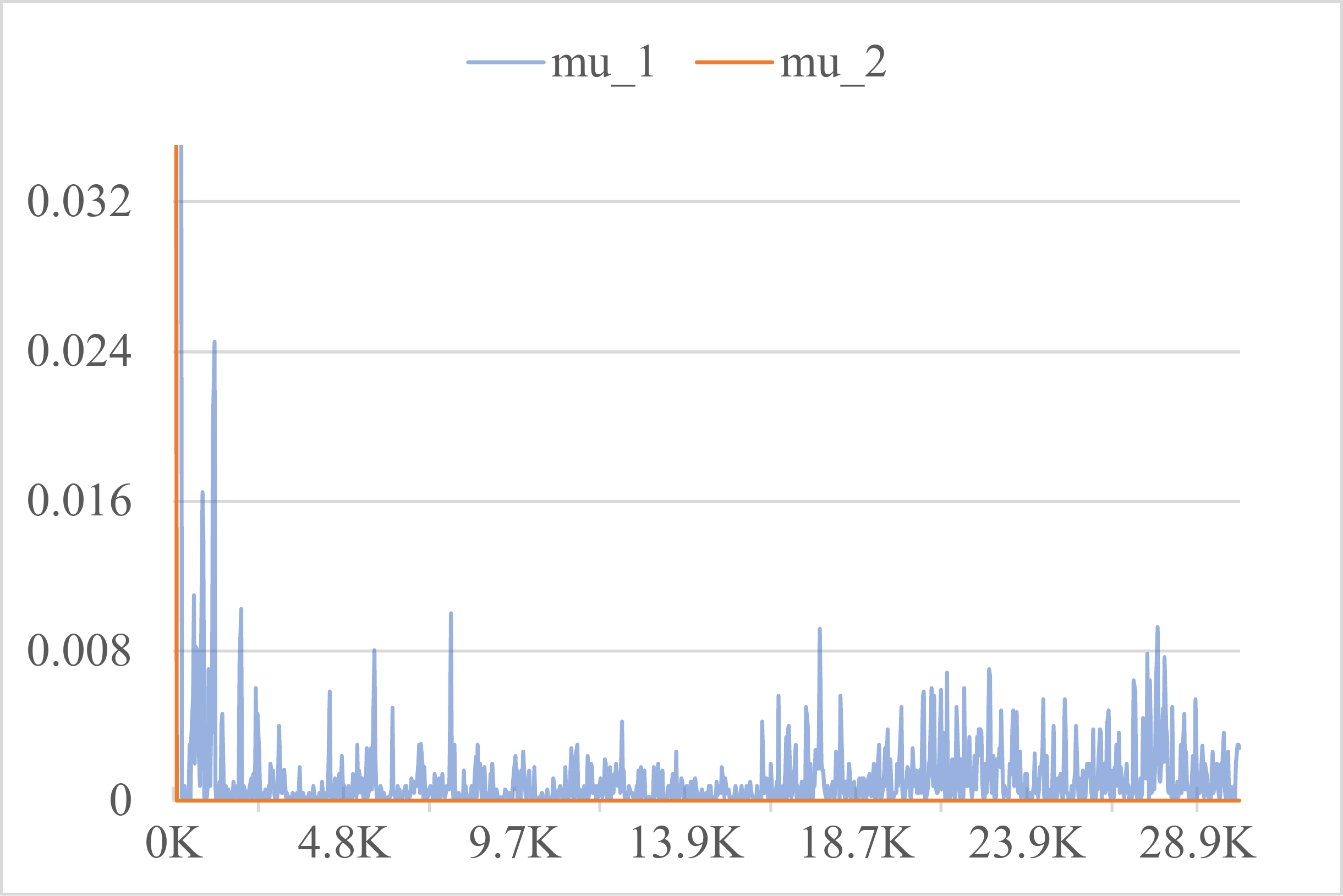}
\caption{\footnotesize FEVEROUS}
\label{fig:feverous_mu}
\end{subfigure}
\begin{subfigure}{0.48\linewidth}
\includegraphics[width=\linewidth]{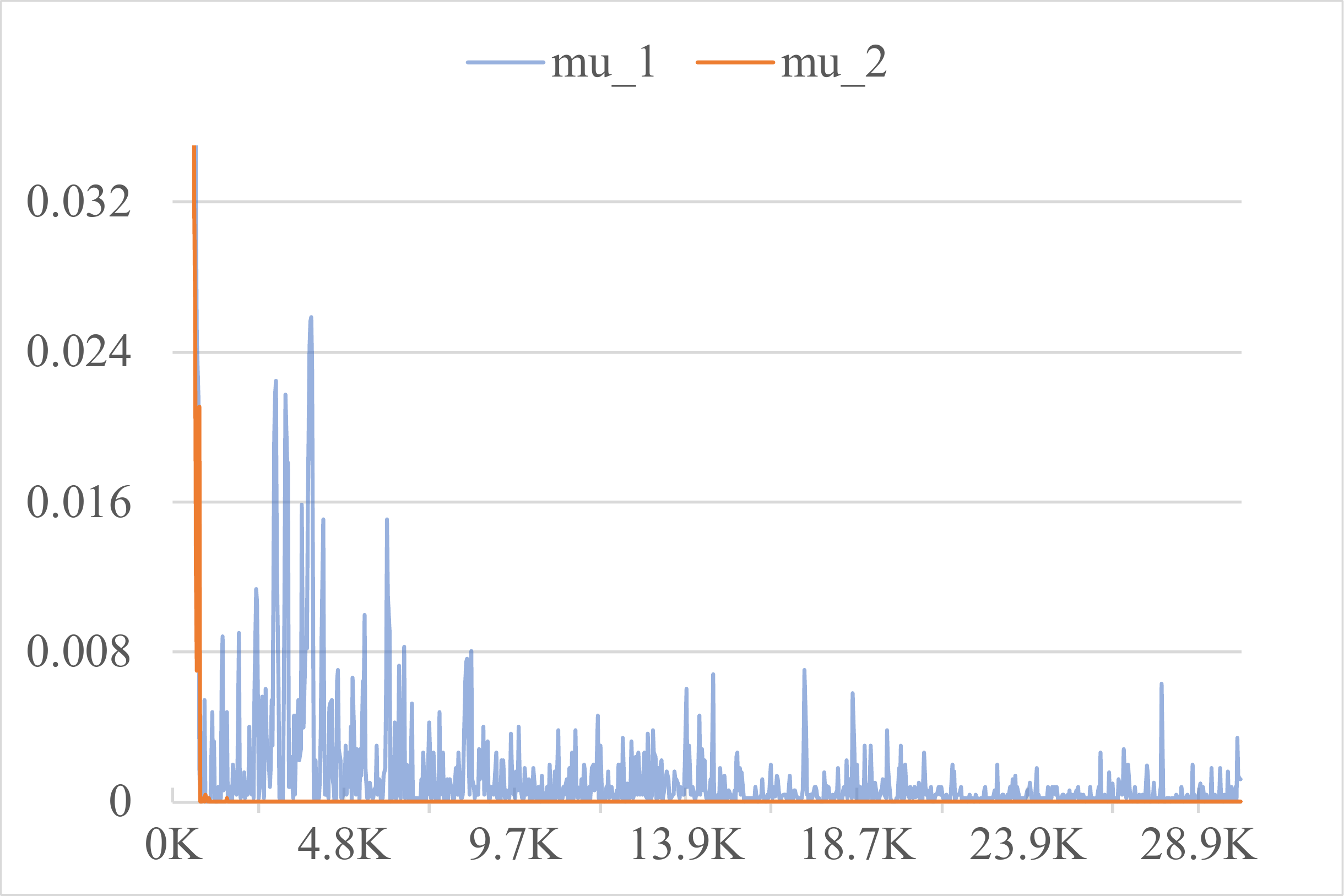}
\caption{\footnotesize HOVER}
\label{fig:hover_mu}
\end{subfigure} 
\caption{The variation of $\mu$ on the FEVEROUS and HOVER during training. The vertical axis represents the scalar value of $\mu$, while the horizontal axis represents the training steps.}
\label{fig:mu}
\end{figure}




To further analyze the superiority of our approach, we conducted a series of ablation experiments, comparing various ablated models with the basic model(basic). Table \ref{tab:main_result} presents the comprehensive performance of our method, demonstrating that each component contributes to the final results in the majority of cases. We also observed a more substantial improvement in performance using our method on the more challenging HOVER. This outcome highlights the potential of our approach in addressing high-difficulty challenges. We then moving forward to analyze the overall effects of each module individually.

\subsubsection{2D Data Augmentation Analysis}

\textbf{Joint training of labels and explaination ensures the explainability}. We first analyzed various settings in 2D Data Augmentation. We first discarded explaination as auxiliary labels(-exp), and directly predicted the first token for classification. We observed that there existing a slight improvement. Next we analyzed whether the model still retained exlainability. Detailed result displayed in Appendix \ref{sec:no_evi} revealed LLMs lost its ability to generate fluent explain and tends to produce repetitive tokens and irrelevant information. Consequently, we consider that sacrificing explainability for marginal performance gains is not a worthwhile trade-off.

\textbf{Both counterfactual dataset and sampling strategy improve performance}. We further evaluate the impact of counterfactual dataset(-cou) on experimental results. Based on our findings, counterfactual dataset did contribute to improve model performance. That's because flipping the truthfulness of facts effectively increased the dataset volume, and the joint fine-tuning of inverted and original datas helped the model effectively learn factual reasoning methods. Finally, we evaluated the impact of the difficulty-based sampling strategy(-sam). The experimental results demonstrates superiority of the sampling strategy in HOVER, which contributed a 6.25\% improvement in f1. We also conducted extensive analysis on specific cases to gain a deeper understanding of the notable effect, and attributed it to the difficulty of the dataset. For a simple data point in FEVEROUS, our model can always generate reasonable explanations given the ground\_true label. However, the same is not true for 3\_hop and 4\_hop examples, as it can only generate reasonable explanations with a probability of 56\% under the same conditions, as shown in Figure \ref{fig:case_study}. The sampling strategy we proposed ensures obtaining more reasonable explanations by repeatedly resampling, counteracting incorrect explanations, and mitigating this issue.



\subsubsection{Improved DPO Analysis}


\textbf{Our improvements to the DPO force model to learn correct reasoning and discard incorrect ones}. In this section, we discuss the impact of our improved DPO. Firstly, we assessed whether the value of $advantage$ affects model performance. We firstly set the advantage to 0(-adv) and observe a significant decline in performance for HOVER and FEVEROUS. Secondly, we test the original DPO(-lag) on the model and also find a noticeable decrease. It is worth noting that both of the settings control the advantage of positive examples over negative examples in terms of generation probability. The former controls how much improvement positive examples should have in generation probability compared to negative examples, while the latter ensures that the generation probability of positive examples is higher than the lower bound of negative examples. These two ablation experiments indicate that our improved DPO, by incorporating the concept of advantage and the idea of comparison, enables the model to learn correct reasoning logic and improves performance.

\textbf{Regularization constraints cannot replace lagrange multipliers}. We further conducted a detailed analysis of $advantage$ in improved DPO. During the training process, we employed the product of $advantage$ and the Lagrange multiplier $\mu$ as a constraint term to regulate the magnitude of the loss, as depicted in Equation \ref{eq:dpo_constrate}. The trainable parameter $\mu$ allows for the scaling of $advantage$. Upon examining the experimental data during training, we observed that $\mu$ frequently scaled $advantage$ to 0, as shown in \ref{fig:mu}, indicating that only a small fraction of the data points resided on the boundary of the constraint. In other words, our improved DPO primarily impacted a select subset of the data points. We conduct further testing on whether setting $\mu$ as a fixed regularization term(-mug) would yield the same experimental results(equivalent to CPO \citep{xu2024contrastive}). Experimental results show a significant decrease in model performance on FEVEROUS and HOVER, especially the latter, with a performance drop of 5.82\% on f1. This indicates that the method of using a regularization term for constraint cannot replace the effectiveness of the improved DPO. The dynamic scaling ability effectively regulates the role of the constraint term, preventing the problem of unstable values and local optima caused by excessive constraint penalties.

\textbf{Learning differences between positive and negative examples effectively improves performance}. Finally, we directly replaced DPO with SFT(-dpo), and experimental results show a significant decrease in performance on both the FEVEROUS and HOVER, particularly the latter, which experienced a performance decrease of 25.1\%. This reveals the role of negative examples in self-instruction, particularly on challenging datasets like HOVER. It is evident that the model can greatly improve its performance by learning the differences between correct and incorrect reasoning paths through contrastive learning.


\subsection{Generalization Performance}

In this section, we assess the model's generalization performance by conducting cross-testing by fine-tuning the model on FEVEROUS, HOVER and an additional dataset called FEVER\citep{Thorne18Fever}. While FEVER and FEVEROUS exhibit some similarities, neither of them presents the same level of difficulty as HOVER. To ensure fairness, all models underwent fine-tuning with counterfactual data. The experimental results shown in Table \ref{tab:OOD}lead us to the following two conclusions.


\textbf{Our proposed method outperforms SFT approaches.} Our method achieves 89\% accuracy and 88.83\% F1 score on FEVEROUS(when fine-tuning the model on HOVER), 73.30\% accuracy and 73.39\% f1 score on HOVER, and 87.22\% accuracy and 87.21\% f1 score on FEVER(when fine-tuning the model on FEVEROUS). We attribute the outstanding performance to 2 factors. On the one hand, compared to traditional methods that focus on label fitting, We utilized a DPO fine-tuning approach with a minimal number of parameters adjusted(trainable parameters accounting for only 0.99\% of the total). This approach preserves the generative and inference capabilities of the language models(LLMs), while full fine-tuning may to some extent compromise the pretraining model's inherent learned knowledge. On the other hand, we propose a method of jointly modeling labels and explanations, which can compel the model to improve its reasoning abilities and generalize them to other datasets.


\textbf{The model exhibits better generalization performance on more challenging datasets(HOVER).} As shown in Table \ref{tab:OOD}, after training on HOVER, our method achieves 89.00\% accuracy and 88.83\% f1 on FEVEROUS, surpassing its performance on HOVER itself. Additionally, on the FEVER dataset, the model achieves an accuracy of 80.94\%. In contrast, when the model is trained on the FEVEROUS dataset, it only achieves 73.30\% accuracy on HOVER. This characteristic can also be observed on other datasets.

\begin{table}
\scriptsize
\begin{tabularx}{\linewidth}{
>{\hsize=0.2\hsize\linewidth=\hsize}X| 
>{\hsize=0.1\hsize\linewidth=\hsize}X
>{\hsize=0.1\hsize\linewidth=\hsize}X 
>{\hsize=0.1\hsize\linewidth=\hsize}X 
>{\hsize=0.1\hsize\linewidth=\hsize}X 
>{\hsize=0.1\hsize\linewidth=\hsize}X 
>{\hsize=0.1\hsize\linewidth=\hsize}X}
\toprule
\multirow{2}{*}{\tb{Dataset}}&\multicolumn{2}{c}{\tb{FEVEROUS}}&\multicolumn{2}{c}{\tb{HOVER}}&\multicolumn{2}{c}{\tb{FEVER}}\\ 
                                    &\bf{ACC}      &\bf{F1}       &\bf{ACC}      &\bf{F1}       &\bf{ACC}      &\bf{F1}        \\ \midrule
\multirow{2}{*}{\tb{OURS}}          &\cg{94.43}    &\cg{94.42}    &\bf{73.30\FIR}&\bf{73.29\FIR}&84.31\SEC    &\bf{84.31\FIR}  \\ 
                                    &\bf{89.00\FIR}&\bf{88.83\FIR}&\cg{80.95}    &\cg{80.43}    & 80.94        & 80.88         \\ \midrule
\multirow{2}{*}{\tb{Bert-FC}}       &\cg{92.27}    &\cg{92.26}    &62.32\THI     & 62.60\THI    & 72.54        & 71.84         \\ 
                                    & 81.39        & 81.31        &\cg{80.27}    &\cg{80.10}    & 60.89        & 60.88         \\ \midrule
\multirow{2}{*}{\tb{LisT5}}         &\cg{93.89}    &\cg{93.87}    &65.05\SEC     &64.68\SEC     &83.27\THI     &83.27\THI      \\ 
                                    &87.60\SEC     &87.43\SEC     &\cg{82.24}    &\cg{82.12}    & 80.00        & 79.75         \\ \midrule
\multirow{2}{*}{\tb{Ro-NLI}}        &\cg{91.63}    &\cg{91.57}    & 60.95        & 60.60        & 60.48        & 54.78         \\ 
                                    &86.22\THI     & 86.00        &\cg{84.42}    &\cg{84.41}    & 77.49        & 77.28         \\ \midrule
\multirow{2}{*}{\tb{De-NLI}}        &\cg{94.00}    &\cg{94.02}    &60.52         & 57.73        &\bf{84.49\FIR}&84.29\SEC      \\ 
                                    & 86.09        &86.06\SEC     &\cg{80.53}    &\cg{80.35}    & 69.66        & 67.67         \\ \midrule
\end{tabularx}

\caption{The performance of model generalization. The first column of the table indicates the dataset on which we conducted the training. We also use \cg{gray font} to indicate the training dataset. Ro-NLI stands for RoBERTa-NLI, and De-NLI stands for DeRoBERTa-NLI.}
\label{tab:OOD}
\end{table}

\subsection{Case Study}

\begin{table*}[t]
\renewcommand{\arraystretch}{1}
{\fontsize{7}{8}\selectfont
\begin{tabularx}{\linewidth}{
>{\hsize=0.1\hsize\linewidth=\hsize}X 
>{\hsize=0.2\hsize\linewidth=\hsize}X 
>{\hsize=0.55\hsize\linewidth=\hsize}X}
\toprule[1.1pt]
\tb{Example 1} & \tb{Label: Refutes} & \tb{Dataset: FEVEROUS} \\\midrule 
\tb{CLAIM} & \multicolumn{2}{X}{Convoy JW 54A(\cred{an tropical convoy}) consisted of 19 merchant ships, close escorts' were provided by destroyers, and it was also accompanied by a local escort group from Britain.} \\ 
\tb{EVIDENCE} & \multicolumn{2}{X}{\tb{Convoy JW 54A}: Convoy JW 54A was an \cred{Arctic convoys} \cg{of World War II sent from United Kingdom by the Allies of World War II to aid the Soviet Union during World War II.}} \\
\tb{ChatGPT} & \multicolumn{2}{X}{Yes, the evidence contradicts the claim. The claim states that Convoy JW 54A was a tropical convoy consisting of 19 merchant ships, with close escorts provided by destroyers and accompanied by a local escort group from Britain. \cred{However, the evidence indicates that Convoy JW 54A was an Arctic convoy} sent from the United Kingdom by the Allies of World War II to aid the Soviet Union. This contradicts the claim regarding the \cred{nature} and \cgreen{location} of the convoy.}\\
\tb{Our Method} & \multicolumn{2}{X}{Yes, the evidence contradicts the claim. According to the evidence, \cred{Convoy JW 54A was an Arctic convoy, not a tropical convoy.} The evidence specifically states that it was an Arctic convoy sent from the United Kingdom by the Allies during World War II to aid the Soviet Union. \cgreen{This contradicts the claim that the convoy consisted of 19 merchant ships, close escorts were provided by destroyers, and it was also accompanied by a local escort group from Britain, as these details are not consistent with an Arctic convoy.}}\\
\toprule[1.1pt]
\tb{Example 2} & \tb{Label: Refutes} & \tb{Dataset: HOVER(3\_hops)} \\\midrule 
\tb{CLAIM} & \multicolumn{2}{X}{The \cred{author of What is Property} was a member of French Parliament. He inspired a political group founded by \cred{Boris Souvarine and Édouard Berth.}} \\ 
\tb{EVIDENCE} & \multicolumn{2}{X}{\cg{1. \tb{Pierre-Joseph Proudhon}: Pierre-Joseph Proudhon(15 January 1809–19 January 1865) was a French politician and the founder of mutualist philosophy. He was the first person to declare himself an anarchist and is widely regarded as one of the ideology's most influential theorists. Proudhon is even considered by many to be the father of anarchism. He became a member of the French Parliament after the revolution of 1848, where after he referred to himself as a federalist.}\newline
2. \tb{Cercle Proudhon}: \cred{\ding{174}Cercle Proudhon(Circle Proudhon) was a political group founded in France on December 16, 1911 by George Valois and Édouard Berth.\ding{173}The group was inspired by Georges Sorel and a selective reading of Proudhon.}\newline
3. \tb{What Is Property?}: or, An Inquiry into the Principle of Right and of Government \cg{(French: Qu'est-ce que la propriété? ou Recherche sur le principe du Droit et du Gouvernement)} is a work of nonfiction on the concept of property and its relation to anarchist philosophy by the French anarchist and \cred{\ding{172}mutualist Pierre-Joseph Proudhon, first published in 1840.} In the book, Proudhon most famously declared that property is theft.} \\
\tb{ChatGPT} & \multicolumn{2}{X}{No, the evidence does not contradict the claim. Pierre-Joseph Proudhon, the author of \"What Is Property?\", was indeed a member of the French Parliament and is known for his anarchist philosophy. The evidence also mentions that the political group Cercle Proudhon was inspired by Georges Sorel, a selective reading of Proudhon's work, and was founded by George Valois, who was associated with Proudhon's ideas. Therefore, the evidence supports the claim.}\\
\tb{Our Method} & \multicolumn{2}{X}{Yes, the evidence contradicts the claim. According to the evidence, \cred{the political group Cercle Proudhon was founded by Georges Valois and Édouard Berth in 1911, and it was inspired by Georges Sorel and a selective reading of Proudhon, but not by Boris Souvarine.} Therefore, the claim that the author of What Is Property was a member of French Parliament and inspired a political group \cred{founded by Boris Souvarine is not supported by the evidence.}}\\
\bottomrule[1.1pt]
\end{tabularx}
}

\caption{Two examples sampled from FEVEROUS and HOVER, along with corresponding response generated by ChatGPT and our method are shown above. We use \cred{red font} and \cred{encircled numbers} to highlight the key information required for the reasoning process, \cg{gray font} to indicate irrelevant information, and \cgreen{green font} to indicate the redundant information generated by the model.}
\label{tab:case_study}
\end{table*}

In this section, we focus on the following questions: What constrains the performance of the model on complex fact-checking tasks? For smaller-scale LLMs, how much improvement does our proposed method offer, and can it provide highly explaination? To answer these questions, we selected one data point from both FEVEROUS and HOVER datasets for analysis, as shown in Table \ref{tab:case_study}. The latter is a 3\_hops fact-checking task with higher inference difficulty. We use \textbf{\textcolor{red}{\ding{172}→\ding{173}→\ding{174}}} to indicate the logical reasoning process.



\textbf{Our method improves LLMs performance but doesn't address the issue of generating illusory information.} From Example 1, we observe that both our proposed method and ChatGPT correctly identify the main contradiction between the claim and evidence(Convoy JW 54A being an Arctic convoy rather than a tropical convoy). However, we also note that the model outputs redundant information, specifically stating that the features of Convoy JW 54A,  such as consisting of 19 merchant ships and being accompanied by a local escort group from Britain, contradict the details of an Arctic convoy. However, the claim and evidence do not explicitly indicate a contradiction between the upon features and it being an Arctic convoy. This is an example of the model generating illusory information, where it attempts to provide additional information to emphasize the contradiction between the claim and evidence. This case also highlights a limitation of our approach in effectively suppressing the generation of illusory information by language models through self-instruction. Addressing this issue should be a focus of future research.



\textbf{Our method can improve the explainability of the model.} As shown in Figure \ref{fig:case_study}, compared to the method before fine-tuning, our method has a significant performance improvement in identifying the main contradictions, with improvements of 14.28\%, 2.50\%, and 11.63\% respectively in 4\_HOP, 3\_HOP, and 2\_HOP, and in most cases it is comparable to ChatGPT. The performance is mostly comparable to ChatGPT, and it can generate longer text content. From the table, we can also observe that using ChatGPT to generate prompts can further improve the model's performance, but this goes against data security, so we did not implement this idea in practice.




\subsection{Manual Evaluation}
\begin{figure}[t]
\centering
\begin{subfigure}{.48\linewidth}
\includegraphics[width=\linewidth]{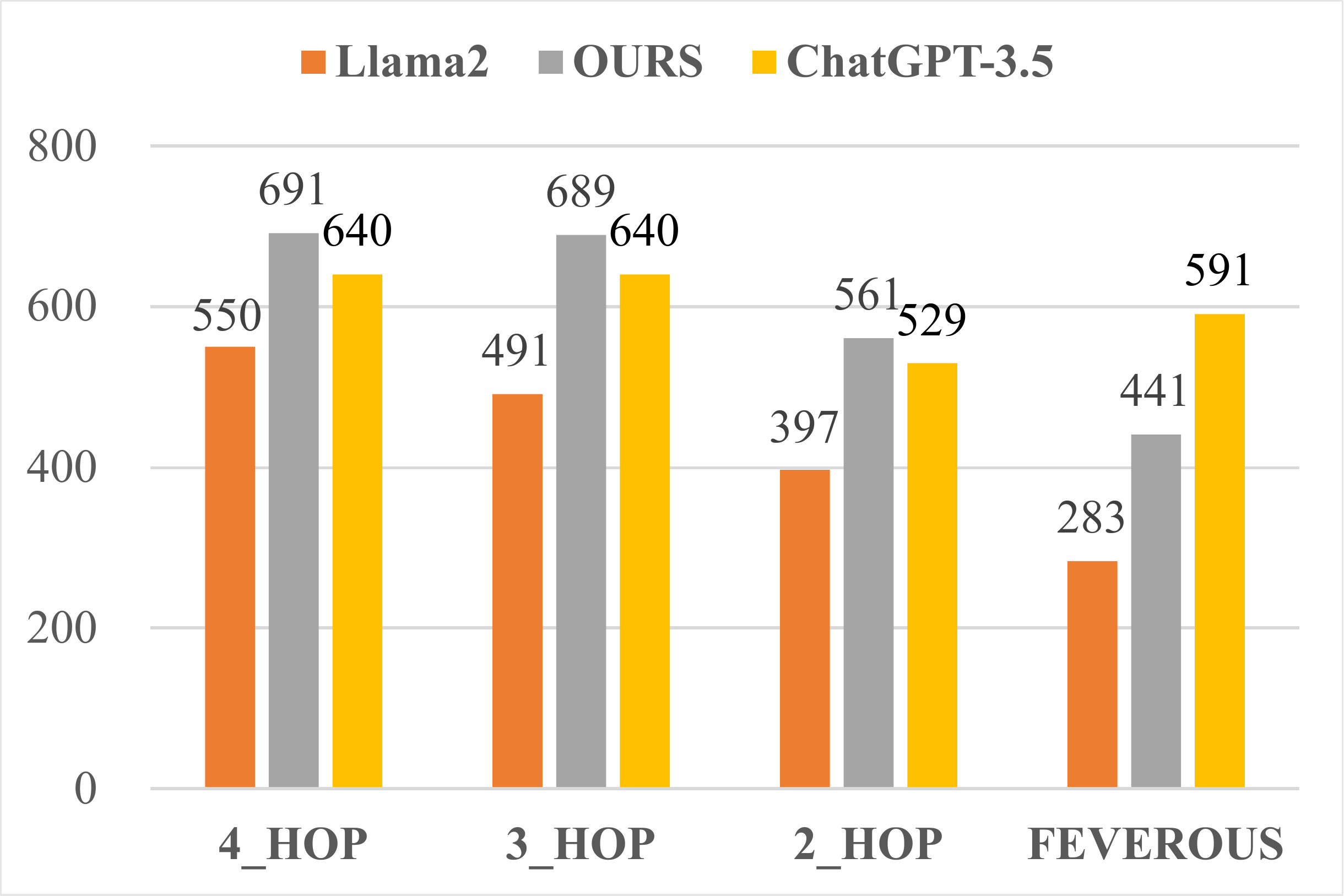}
\caption{Text length.}
\label{fig:text_length}
\end{subfigure}
\begin{subfigure}{.48\linewidth}
\includegraphics[width=\linewidth]{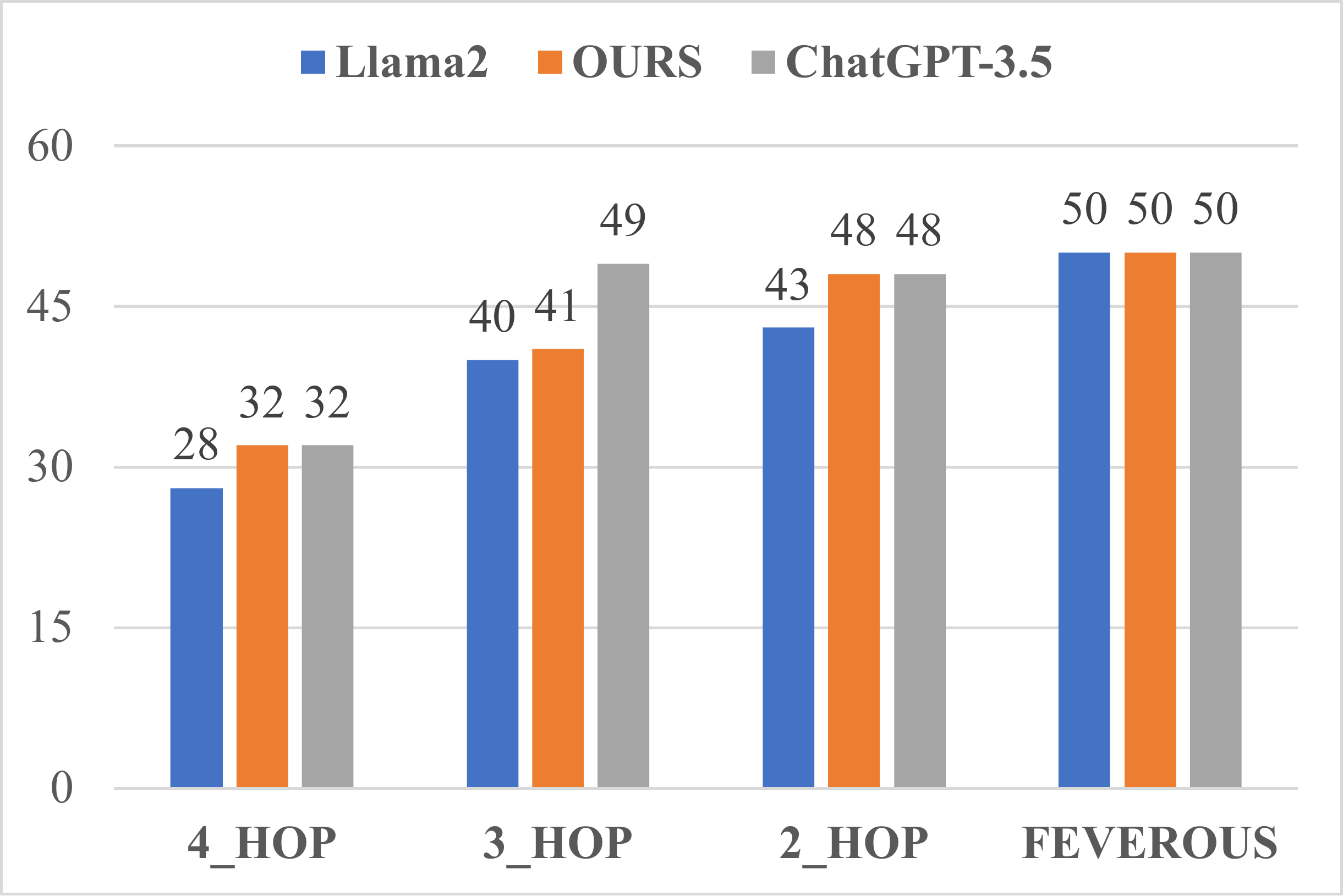}
\caption{Explanation Counting.}
\label{fig:explaination_counter}
\end{subfigure}
\caption{The length of the generated explaination and the count of reasonable explanations. The horizontal axis represents the data type, and the vertical axis represents the text length(\ref{fig:text_length}) and the count of reasonable explanations( \ref{fig:explaination_counter}).}
\label{fig:case_study}
\end{figure}

In this section, we further evaluate the explainability of the models. We adopt a human evaluation to assess whether the models generate reasonable explanations to support the given labels. Therefore, we set up the following experiment: we select 5 samples from the 2\_HOP, 3\_HOP, and 4\_HOP subsets of HOVER, as well as 5 random samples from FEVEROUS. For each sample, we generate 10 explanations given correct labels. We conducted experiment performance of the Llama2, our method, and ChatGPT separately and hired professional annotators to judge whether the generated explanations reveal the main contradictions between claims and the evidence.



\textbf{Our Fine-Tuning Method Enhances Model Explainability}. As shown in Figure \ref{fig:case_study}, our fine-tuning method demonstrates significant performance improvements compared to the pre-fine-tuned model. Specifically, it achieved 14.28\%, 2.50\%, and 11.63\% increases in identifying the main contradictions for 4\_HOP, 3\_HOP, and 2\_HOP respectively, and in most cases, the performance is comparable to ChatGPT. While we observed that using ChatGPT to generate explain could further enhance model performance, we decided not to pursue this approach as it would compromise the security of the data.

\section{Conclusion and Future Work}
Our work attempts to improve model performance using self-instruction and employs an enhanced dpo approach that combines fine-tuning the model with positive and negative samples. We also conducted a series of experiments to validate the effectiveness of our model. First, the overall experimental results demonstrate the accuracy and effectiveness of our approach. Then the ablation study to individually validate the efficacy of each module. Finally, Generalization performance evaluation, case study and manual evaluation were conducted to verify the model's generalization ability and explainability. In future work, we hope to implement the following aspects:

\textbf{Deeper exploration of negative samples in the context of explainable fact-checking.} Our experimental results shows that negative samples contribute to model performance to some extends, and we believe that further investigating the role of negative samples can enhance the performance of LLMs and can be utilized in prompt techniques. Current prompt techniques often employ the idea of COT\citep{10.5555/3600270.3602070} and RAG\citep{lewis2020retrieval}, while overlooking the importance of learning from negative/incorrect samples.


\textbf{More comprehensive self-instruction approach.} Our experiments only provided a simple self-instruction scheme while preserving model interpretability. Due to the limitations of open-source LLMs, the model may struggle to generate reliable explanatory content for more complex fact-checking tasks in order to achieve self-instruction. We believe that models like ChatGPT can be utilized to generate interpretable content and facilitate model distillation. When dealing with multi-hop fact-checking tasks, we can fine-tune the model in multiple steps, dividing the task into several stages.


\textbf{Different Experiment in an open-world setting.} Our work focuses on improving the performance of LLMs given complex evidence, which often contains a large amount of irrelevant information. However, real-world fact-checking tasks involve more complex processes, such as information retrieval. In future work, we aim to apply our method to fact-checking and debunking fake news in an open-world scenario.

\section{Limitation}

As mentioned in Section \ref{sec:result_and_discussion}, our method has some limitations, which we will now summarize.

Firstly, our method does not address the challenge of suppressing the generation of hallucinatory information by LLMs. As a result, the generated outputs may occasionally contain misleading information, even if the predictions themselves are accurate.

Secondly, although we have carefully designed the training strategy for our model, the self-instruction mechanism itself is relatively simple. This means that the explanatory content generated by the model is entirely controlled by LLMs, leading to a random arrangement of information and hindering the systematic analysis of the generated content by fact-checkers.

Lastly, due to the performance limitations of Llama2, the model may struggle to generate accurate explanations when confronted with complex multi-hop evidence. This is one of the reasons why our method falls short of surpassing traditional methods in terms of accuracy. To address this issue, we aim to mitigate the problem by leveraging the capabilities of commercial models through distillation.

\bibliographystyle{plainnat}
\bibliography{references.bib}

\begin{thebibliography}{47}
\providecommand{\natexlab}[1]{#1}
\providecommand{\url}[1]{\texttt{#1}}
\expandafter\ifx\csname urlstyle\endcsname\relax
  \providecommand{\doi}[1]{doi: #1}\else
  \providecommand{\doi}{doi: \begingroup \urlstyle{rm}\Url}\fi

\bibitem[Aly et~al.(2021)Aly, Guo, Schlichtkrull, Thorne, Vlachos, Christodoulopoulos, Cocarascu, and Mittal]{aly-etal-2021-fact}
Rami Aly, Zhijiang Guo, Michael~Sejr Schlichtkrull, James Thorne, Andreas Vlachos, Christos Christodoulopoulos, Oana Cocarascu, and Arpit Mittal.
\newblock The fact extraction and {VER}ification over unstructured and structured information ({FEVEROUS}) shared task.
\newblock In Rami Aly, Christos Christodoulopoulos, Oana Cocarascu, Zhijiang Guo, Arpit Mittal, Michael Schlichtkrull, James Thorne, and Andreas Vlachos, editors, \emph{Proceedings of the Fourth Workshop on Fact Extraction and VERification (FEVER)}, pages 1--13, Dominican Republic, November 2021. Association for Computational Linguistics.
\newblock \doi{10.18653/v1/2021.fever-1.1}.
\newblock URL \url{https://aclanthology.org/2021.fever-1.1}.

\bibitem[Bai et~al.(2022)Bai, Jones, Ndousse, Askell, Chen, DasSarma, Drain, Fort, Ganguli, Henighan, Joseph, Kadavath, Kernion, Conerly, El-Showk, Elhage, Hatfield-Dodds, Hernandez, Hume, Johnston, Kravec, Lovitt, Nanda, Olsson, Amodei, Brown, Clark, McCandlish, Olah, Mann, and Kaplan]{bai2022training}
Yuntao Bai, Andy Jones, Kamal Ndousse, Amanda Askell, Anna Chen, Nova DasSarma, Dawn Drain, Stanislav Fort, Deep Ganguli, Tom Henighan, Nicholas Joseph, Saurav Kadavath, Jackson Kernion, Tom Conerly, Sheer El-Showk, Nelson Elhage, Zac Hatfield-Dodds, Danny Hernandez, Tristan Hume, Scott Johnston, Shauna Kravec, Liane Lovitt, Neel Nanda, Catherine Olsson, Dario Amodei, Tom Brown, Jack Clark, Sam McCandlish, Chris Olah, Ben Mann, and Jared Kaplan.
\newblock Training a helpful and harmless assistant with reinforcement learning from human feedback, 2022.

\bibitem[Bertsekas(2014)]{bertsekas2014constrained}
Dimitri~P Bertsekas.
\newblock \emph{Constrained optimization and Lagrange multiplier methods}.
\newblock Academic press, 2014.

\bibitem[Botnevik et~al.(2020)Botnevik, Sakariassen, and Setty]{10.1145/3397271.3401396}
Bjarte Botnevik, Eirik Sakariassen, and Vinay Setty.
\newblock Brenda: Browser extension for fake news detection.
\newblock In \emph{Proceedings of the 43rd International ACM SIGIR Conference on Research and Development in Information Retrieval}, SIGIR '20, page 2117–2120, New York, NY, USA, 2020. Association for Computing Machinery.
\newblock ISBN 9781450380164.
\newblock \doi{10.1145/3397271.3401396}.
\newblock URL \url{https://doi.org/10.1145/3397271.3401396}.

\bibitem[Brown et~al.(2020)Brown, Mann, Ryder, Subbiah, Kaplan, Dhariwal, Neelakantan, Shyam, Sastry, Askell, Agarwal, Herbert-Voss, Krueger, Henighan, Child, Ramesh, Ziegler, Wu, Winter, Hesse, Chen, Sigler, Litwin, Gray, Chess, Clark, Berner, McCandlish, Radford, Sutskever, and Amodei]{10.5555/3495724.3495883}
Tom~B. Brown, Benjamin Mann, Nick Ryder, Melanie Subbiah, Jared Kaplan, Prafulla Dhariwal, Arvind Neelakantan, Pranav Shyam, Girish Sastry, Amanda Askell, Sandhini Agarwal, Ariel Herbert-Voss, Gretchen Krueger, Tom Henighan, Rewon Child, Aditya Ramesh, Daniel~M. Ziegler, Jeffrey Wu, Clemens Winter, Christopher Hesse, Mark Chen, Eric Sigler, Mateusz Litwin, Scott Gray, Benjamin Chess, Jack Clark, Christopher Berner, Sam McCandlish, Alec Radford, Ilya Sutskever, and Dario Amodei.
\newblock Language models are few-shot learners.
\newblock In \emph{Proceedings of the 34th International Conference on Neural Information Processing Systems}, NIPS '20, Red Hook, NY, USA, 2020. Curran Associates Inc.
\newblock ISBN 9781713829546.

\bibitem[Chen et~al.(2023)Chen, Kim, Sriram, Durrett, and Choi]{chen2023complex}
Jifan Chen, Grace Kim, Aniruddh Sriram, Greg Durrett, and Eunsol Choi.
\newblock Complex claim verification with evidence retrieved in the wild, 2023.

\bibitem[Chiang et~al.(2022)Chiang, Chuang, and Lee]{chiang-etal-2022-recent}
Cheng-Han Chiang, Yung-Sung Chuang, and Hung-yi Lee.
\newblock Recent advances in pre-trained language models: Why do they work and how do they work.
\newblock In Miguel~A. Alonso and Zhongyu Wei, editors, \emph{Proceedings of the 2nd Conference of the Asia-Pacific Chapter of the Association for Computational Linguistics and the 12th International Joint Conference on Natural Language Processing: Tutorial Abstracts}, pages 8--15, Taipei, November 2022. Association for Computational Linguistics.
\newblock URL \url{https://aclanthology.org/2022.aacl-tutorials.2}.

\bibitem[Chung et~al.(2024)Chung, Hou, Longpre, Zoph, Tay, Fedus, Li, Wang, Dehghani, Brahma, et~al.]{chung2024scaling}
Hyung~Won Chung, Le~Hou, Shayne Longpre, Barret Zoph, Yi~Tay, William Fedus, Yunxuan Li, Xuezhi Wang, Mostafa Dehghani, Siddhartha Brahma, et~al.
\newblock Scaling instruction-finetuned language models.
\newblock \emph{Journal of Machine Learning Research}, 25\penalty0 (70):\penalty0 1--53, 2024.

\bibitem[Das et~al.(2023)Das, Liu, Kovatchev, and Lease]{DAS2023103219}
Anubrata Das, Houjiang Liu, Venelin Kovatchev, and Matthew Lease.
\newblock The state of human-centered nlp technology for fact-checking.
\newblock \emph{Information Processing \& Management}, 60\penalty0 (2):\penalty0 103219, 2023.
\newblock ISSN 0306-4573.
\newblock \doi{https://doi.org/10.1016/j.ipm.2022.103219}.
\newblock URL \url{https://www.sciencedirect.com/science/article/pii/S030645732200320X}.

\bibitem[Glockner et~al.(2022)Glockner, Hou, and Gurevych]{glockner-etal-2022-missing}
Max Glockner, Yufang Hou, and Iryna Gurevych.
\newblock Missing counter-evidence renders {NLP} fact-checking unrealistic for misinformation.
\newblock In Yoav Goldberg, Zornitsa Kozareva, and Yue Zhang, editors, \emph{Proceedings of the 2022 Conference on Empirical Methods in Natural Language Processing}, pages 5916--5936, Abu Dhabi, United Arab Emirates, December 2022. Association for Computational Linguistics.
\newblock \doi{10.18653/v1/2022.emnlp-main.397}.
\newblock URL \url{https://aclanthology.org/2022.emnlp-main.397}.

\bibitem[Guo et~al.(2022)Guo, Schlichtkrull, and Vlachos]{10.1162/tacl_a_00454}
Zhijiang Guo, Michael Schlichtkrull, and Andreas Vlachos.
\newblock {A Survey on Automated Fact-Checking}.
\newblock \emph{Transactions of the Association for Computational Linguistics}, 10:\penalty0 178--206, 02 2022.
\newblock ISSN 2307-387X.
\newblock \doi{10.1162/tacl_a_00454}.
\newblock URL \url{https://doi.org/10.1162/tacl\_a\_00454}.

\bibitem[Hu et~al.(2024)Hu, Sheng, Cao, Shi, Li, Wang, and Qi]{Hu_Sheng_Cao_Shi_Li_Wang_Qi_2024}
Beizhe Hu, Qiang Sheng, Juan Cao, Yuhui Shi, Yang Li, Danding Wang, and Peng Qi.
\newblock Bad actor, good advisor: Exploring the role of large language models in fake news detection.
\newblock \emph{Proceedings of the AAAI Conference on Artificial Intelligence}, 38\penalty0 (20):\penalty0 22105--22113, Mar. 2024.
\newblock \doi{10.1609/aaai.v38i20.30214}.
\newblock URL \url{https://ojs.aaai.org/index.php/AAAI/article/view/30214}.

\bibitem[Hu et~al.(2022)Hu, Shen, Wallis, Allen-Zhu, Li, Wang, Wang, and Chen]{hu2022lora}
Edward~J Hu, Yelong Shen, Phillip Wallis, Zeyuan Allen-Zhu, Yuanzhi Li, Shean Wang, Lu~Wang, and Weizhu Chen.
\newblock Lo{RA}: Low-rank adaptation of large language models.
\newblock In \emph{International Conference on Learning Representations}, 2022.
\newblock URL \url{https://openreview.net/forum?id=nZeVKeeFYf9}.

\bibitem[Hu et~al.(2023)Hu, Hong, Guo, Wen, and Yu]{10.1145/3539618.3592049}
Xuming Hu, Zhaochen Hong, Zhijiang Guo, Lijie Wen, and Philip Yu.
\newblock Read it twice: Towards faithfully interpretable fact verification by revisiting evidence.
\newblock In \emph{Proceedings of the 46th International ACM SIGIR Conference on Research and Development in Information Retrieval}, SIGIR '23, page 2319–2323, New York, NY, USA, 2023. Association for Computing Machinery.
\newblock ISBN 9781450394086.
\newblock \doi{10.1145/3539618.3592049}.
\newblock URL \url{https://doi.org/10.1145/3539618.3592049}.

\bibitem[Jiang et~al.(2021)Jiang, Pradeep, and Lin]{jiang-etal-2021-exploring-listwise}
Kelvin Jiang, Ronak Pradeep, and Jimmy Lin.
\newblock Exploring listwise evidence reasoning with t5 for fact verification.
\newblock In Chengqing Zong, Fei Xia, Wenjie Li, and Roberto Navigli, editors, \emph{Proceedings of the 59th Annual Meeting of the Association for Computational Linguistics and the 11th International Joint Conference on Natural Language Processing (Volume 2: Short Papers)}, pages 402--410, Online, August 2021. Association for Computational Linguistics.
\newblock \doi{10.18653/v1/2021.acl-short.51}.
\newblock URL \url{https://aclanthology.org/2021.acl-short.51}.

\bibitem[Lewis et~al.(2020)Lewis, Perez, Piktus, Petroni, Karpukhin, Goyal, K{\"u}ttler, Lewis, Yih, Rockt{\"a}schel, et~al.]{lewis2020retrieval}
Patrick Lewis, Ethan Perez, Aleksandra Piktus, Fabio Petroni, Vladimir Karpukhin, Naman Goyal, Heinrich K{\"u}ttler, Mike Lewis, Wen-tau Yih, Tim Rockt{\"a}schel, et~al.
\newblock Retrieval-augmented generation for knowledge-intensive nlp tasks.
\newblock \emph{Advances in Neural Information Processing Systems}, 33:\penalty0 9459--9474, 2020.

\bibitem[Li et~al.(2023)Li, Peng, and Zhang]{Li2023SelfCheckerPM}
Miaoran Li, Baolin Peng, and Zhu Zhang.
\newblock Self-checker: Plug-and-play modules for fact-checking with large language models.
\newblock \emph{ArXiv}, abs/2305.14623, 2023.
\newblock URL \url{https://api.semanticscholar.org/CorpusID:258865801}.

\bibitem[Ma et~al.(2023)Ma, Cao, Hong, and Sun]{ma-etal-2023-large}
Yubo Ma, Yixin Cao, Yong Hong, and Aixin Sun.
\newblock Large language model is not a good few-shot information extractor, but a good reranker for hard samples!
\newblock In Houda Bouamor, Juan Pino, and Kalika Bali, editors, \emph{Findings of the Association for Computational Linguistics: EMNLP 2023}, pages 10572--10601, Singapore, December 2023. Association for Computational Linguistics.
\newblock \doi{10.18653/v1/2023.findings-emnlp.710}.
\newblock URL \url{https://aclanthology.org/2023.findings-emnlp.710}.

\bibitem[Ma et~al.(2022)Ma, Li, Li, and Cheng]{ma-etal-2022-glaf}
Zhiyuan Ma, Jianjun Li, Guohui Li, and Yongjing Cheng.
\newblock {GLAF}: Global-to-local aggregation and fission network for semantic level fact verification.
\newblock In Nicoletta Calzolari, Chu-Ren Huang, Hansaem Kim, James Pustejovsky, Leo Wanner, Key-Sun Choi, Pum-Mo Ryu, Hsin-Hsi Chen, Lucia Donatelli, Heng Ji, Sadao Kurohashi, Patrizia Paggio, Nianwen Xue, Seokhwan Kim, Younggyun Hahm, Zhong He, Tony~Kyungil Lee, Enrico Santus, Francis Bond, and Seung-Hoon Na, editors, \emph{Proceedings of the 29th International Conference on Computational Linguistics}, pages 1801--1812, Gyeongju, Republic of Korea, October 2022. International Committee on Computational Linguistics.
\newblock URL \url{https://aclanthology.org/2022.coling-1.155}.

\bibitem[Mangrulkar et~al.(2022)Mangrulkar, Gugger, Debut, Belkada, Paul, and Bossan]{peft}
Sourab Mangrulkar, Sylvain Gugger, Lysandre Debut, Younes Belkada, Sayak Paul, and Benjamin Bossan.
\newblock Peft: State-of-the-art parameter-efficient fine-tuning methods.
\newblock \url{https://github.com/huggingface/peft}, 2022.

\bibitem[Naveed et~al.(2023)Naveed, Khan, Qiu, Saqib, Anwar, Usman, Barnes, and Mian]{Naveed2023ACO}
Humza Naveed, Asad~Ullah Khan, Shi Qiu, Muhammad Saqib, Saeed Anwar, Muhammad Usman, Nick Barnes, and Ajmal~S. Mian.
\newblock A comprehensive overview of large language models.
\newblock \emph{ArXiv}, abs/2307.06435, 2023.
\newblock URL \url{https://api.semanticscholar.org/CorpusID:259847443}.

\bibitem[Nie et~al.(2020)Nie, Williams, Dinan, Bansal, Weston, and Kiela]{nie-etal-2020-adversarial}
Yixin Nie, Adina Williams, Emily Dinan, Mohit Bansal, Jason Weston, and Douwe Kiela.
\newblock Adversarial {NLI}: A new benchmark for natural language understanding.
\newblock In Dan Jurafsky, Joyce Chai, Natalie Schluter, and Joel Tetreault, editors, \emph{Proceedings of the 58th Annual Meeting of the Association for Computational Linguistics}, pages 4885--4901, Online, July 2020. Association for Computational Linguistics.
\newblock \doi{10.18653/v1/2020.acl-main.441}.
\newblock URL \url{https://aclanthology.org/2020.acl-main.441}.

\bibitem[Ouyang et~al.(2022)Ouyang, Wu, Jiang, Almeida, Wainwright, Mishkin, Zhang, Agarwal, Slama, Ray, Schulman, Hilton, Kelton, Miller, Simens, Askell, Welinder, Christiano, Leike, and Lowe]{ouyang2022training}
Long Ouyang, Jeff Wu, Xu~Jiang, Diogo Almeida, Carroll~L. Wainwright, Pamela Mishkin, Chong Zhang, Sandhini Agarwal, Katarina Slama, Alex Ray, John Schulman, Jacob Hilton, Fraser Kelton, Luke Miller, Maddie Simens, Amanda Askell, Peter Welinder, Paul Christiano, Jan Leike, and Ryan Lowe.
\newblock Training language models to follow instructions with human feedback, 2022.

\bibitem[Pan et~al.(2023)Pan, Wu, Lu, Luu, Wang, Kan, and Nakov]{pan-etal-2023-fact}
Liangming Pan, Xiaobao Wu, Xinyuan Lu, Anh~Tuan Luu, William~Yang Wang, Min-Yen Kan, and Preslav Nakov.
\newblock Fact-checking complex claims with program-guided reasoning.
\newblock In Anna Rogers, Jordan Boyd-Graber, and Naoaki Okazaki, editors, \emph{Proceedings of the 61st Annual Meeting of the Association for Computational Linguistics (Volume 1: Long Papers)}, pages 6981--7004, Toronto, Canada, July 2023. Association for Computational Linguistics.
\newblock \doi{10.18653/v1/2023.acl-long.386}.
\newblock URL \url{https://aclanthology.org/2023.acl-long.386}.

\bibitem[Rafailov et~al.(2023)Rafailov, Sharma, Mitchell, Manning, Ermon, and Finn]{rafailov2023direct}
Rafael Rafailov, Archit Sharma, Eric Mitchell, Christopher~D Manning, Stefano Ermon, and Chelsea Finn.
\newblock Direct preference optimization: Your language model is secretly a reward model.
\newblock In \emph{Thirty-seventh Conference on Neural Information Processing Systems}, 2023.
\newblock URL \url{https://openreview.net/forum?id=HPuSIXJaa9}.

\bibitem[Sheng et~al.(2021)Sheng, Cao, Zhang, Li, and Zhong]{sheng-etal-2021-article}
Qiang Sheng, Juan Cao, Xueyao Zhang, Xirong Li, and Lei Zhong.
\newblock Article reranking by memory-enhanced key sentence matching for detecting previously fact-checked claims.
\newblock In Chengqing Zong, Fei Xia, Wenjie Li, and Roberto Navigli, editors, \emph{Proceedings of the 59th Annual Meeting of the Association for Computational Linguistics and the 11th International Joint Conference on Natural Language Processing (Volume 1: Long Papers)}, pages 5468--5481, Online, August 2021. Association for Computational Linguistics.
\newblock \doi{10.18653/v1/2021.acl-long.425}.
\newblock URL \url{https://aclanthology.org/2021.acl-long.425}.

\bibitem[Shu et~al.(2019)Shu, Cui, Wang, Lee, and Liu]{10.1145/3292500.3330935}
Kai Shu, Limeng Cui, Suhang Wang, Dongwon Lee, and Huan Liu.
\newblock defend: Explainable fake news detection.
\newblock In \emph{Proceedings of the 25th ACM SIGKDD International Conference on Knowledge Discovery \& Data Mining}, KDD '19, page 395–405, New York, NY, USA, 2019. Association for Computing Machinery.
\newblock ISBN 9781450362016.
\newblock \doi{10.1145/3292500.3330935}.
\newblock URL \url{https://doi.org/10.1145/3292500.3330935}.

\bibitem[Si et~al.(2021)Si, Zhou, Li, Shi, and He]{si-etal-2021-topic}
Jiasheng Si, Deyu Zhou, Tongzhe Li, Xingyu Shi, and Yulan He.
\newblock Topic-aware evidence reasoning and stance-aware aggregation for fact verification.
\newblock In Chengqing Zong, Fei Xia, Wenjie Li, and Roberto Navigli, editors, \emph{Proceedings of the 59th Annual Meeting of the Association for Computational Linguistics and the 11th International Joint Conference on Natural Language Processing (Volume 1: Long Papers)}, pages 1612--1622, Online, August 2021. Association for Computational Linguistics.
\newblock \doi{10.18653/v1/2021.acl-long.128}.
\newblock URL \url{https://aclanthology.org/2021.acl-long.128}.

\bibitem[Si et~al.(2023)Si, Zhu, and Zhou]{10.1609/aaai.v37i11.26591}
Jiasheng Si, Yingjie Zhu, and Deyu Zhou.
\newblock Exploring faithful rationale for multi-hop fact verification via salience-aware graph learning.
\newblock In \emph{Proceedings of the Thirty-Seventh AAAI Conference on Artificial Intelligence and Thirty-Fifth Conference on Innovative Applications of Artificial Intelligence and Thirteenth Symposium on Educational Advances in Artificial Intelligence}, AAAI'23/IAAI'23/EAAI'23. AAAI Press, 2023.
\newblock ISBN 978-1-57735-880-0.
\newblock \doi{10.1609/aaai.v37i11.26591}.
\newblock URL \url{https://doi.org/10.1609/aaai.v37i11.26591}.

\bibitem[Sileo(2023)]{sileo2023tasksource}
Damien Sileo.
\newblock tasksource: Structured dataset preprocessing annotations for frictionless extreme multi-task learning and evaluation.
\newblock \emph{arXiv preprint arXiv:2301.05948}, 2023.
\newblock URL \url{https://arxiv.org/abs/2301.05948}.

\bibitem[Soleimani et~al.(2020)Soleimani, Monz, and Worring]{10.1007/978-3-030-45442-5_45}
Amir Soleimani, Christof Monz, and Marcel Worring.
\newblock Bert for evidence retrieval and claim verification.
\newblock In \emph{Advances in Information Retrieval: 42nd European Conference on IR Research, ECIR 2020, Lisbon, Portugal, April 14–17, 2020, Proceedings, Part II}, page 359–366, Berlin, Heidelberg, 2020. Springer-Verlag.
\newblock ISBN 978-3-030-45441-8.
\newblock \doi{10.1007/978-3-030-45442-5_45}.
\newblock URL \url{https://doi.org/10.1007/978-3-030-45442-5_45}.

\bibitem[Thorne et~al.(2018)Thorne, Vlachos, Christodoulopoulos, and Mittal]{Thorne18Fever}
James Thorne, Andreas Vlachos, Christos Christodoulopoulos, and Arpit Mittal.
\newblock {FEVER}: a large-scale dataset for fact extraction and {VERification}.
\newblock In \emph{NAACL-HLT}, 2018.

\bibitem[Touvron et~al.(2023)Touvron, Martin, Stone, Albert, Almahairi, Babaei, Bashlykov, Batra, Bhargava, Bhosale, Bikel, Blecher, Ferrer, Chen, Cucurull, Esiobu, Fernandes, Fu, Fu, Fuller, Gao, Goswami, Goyal, Hartshorn, Hosseini, Hou, Inan, Kardas, Kerkez, Khabsa, Kloumann, Korenev, Koura, Lachaux, Lavril, Lee, Liskovich, Lu, Mao, Martinet, Mihaylov, Mishra, Molybog, Nie, Poulton, Reizenstein, Rungta, Saladi, Schelten, Silva, Smith, Subramanian, Tan, Tang, Taylor, Williams, Kuan, Xu, Yan, Zarov, Zhang, Fan, Kambadur, Narang, Rodriguez, Stojnic, Edunov, and Scialom]{touvron2023llama}
Hugo Touvron, Louis Martin, Kevin Stone, Peter Albert, Amjad Almahairi, Yasmine Babaei, Nikolay Bashlykov, Soumya Batra, Prajjwal Bhargava, Shruti Bhosale, Dan Bikel, Lukas Blecher, Cristian~Canton Ferrer, Moya Chen, Guillem Cucurull, David Esiobu, Jude Fernandes, Jeremy Fu, Wenyin Fu, Brian Fuller, Cynthia Gao, Vedanuj Goswami, Naman Goyal, Anthony Hartshorn, Saghar Hosseini, Rui Hou, Hakan Inan, Marcin Kardas, Viktor Kerkez, Madian Khabsa, Isabel Kloumann, Artem Korenev, Punit~Singh Koura, Marie-Anne Lachaux, Thibaut Lavril, Jenya Lee, Diana Liskovich, Yinghai Lu, Yuning Mao, Xavier Martinet, Todor Mihaylov, Pushkar Mishra, Igor Molybog, Yixin Nie, Andrew Poulton, Jeremy Reizenstein, Rashi Rungta, Kalyan Saladi, Alan Schelten, Ruan Silva, Eric~Michael Smith, Ranjan Subramanian, Xiaoqing~Ellen Tan, Binh Tang, Ross Taylor, Adina Williams, Jian~Xiang Kuan, Puxin Xu, Zheng Yan, Iliyan Zarov, Yuchen Zhang, Angela Fan, Melanie Kambadur, Sharan Narang, Aurelien Rodriguez, Robert Stojnic, Sergey Edunov, and Thomas
  Scialom.
\newblock Llama 2: Open foundation and fine-tuned chat models, 2023.

\bibitem[von Werra et~al.(2020)von Werra, Belkada, Tunstall, Beeching, Thrush, Lambert, and Huang]{vonwerra2022trl}
Leandro von Werra, Younes Belkada, Lewis Tunstall, Edward Beeching, Tristan Thrush, Nathan Lambert, and Shengyi Huang.
\newblock Trl: Transformer reinforcement learning.
\newblock \url{https://github.com/huggingface/trl}, 2020.

\bibitem[Vosoughi et~al.(2018)Vosoughi, Roy, and Aral]{doi:10.1126/science.aap9559}
Soroush Vosoughi, Deb Roy, and Sinan Aral.
\newblock The spread of true and false news online.
\newblock \emph{Science}, 359\penalty0 (6380):\penalty0 1146--1151, 2018.
\newblock \doi{10.1126/science.aap9559}.
\newblock URL \url{https://www.science.org/doi/abs/10.1126/science.aap9559}.

\bibitem[Wang and Shu(2023)]{wang-shu-2023-explainable}
Haoran Wang and Kai Shu.
\newblock Explainable claim verification via knowledge-grounded reasoning with large language models.
\newblock In Houda Bouamor, Juan Pino, and Kalika Bali, editors, \emph{Findings of the Association for Computational Linguistics: EMNLP 2023}, pages 6288--6304, Singapore, December 2023. Association for Computational Linguistics.
\newblock \doi{10.18653/v1/2023.findings-emnlp.416}.
\newblock URL \url{https://aclanthology.org/2023.findings-emnlp.416}.

\bibitem[Wang et~al.(2022)Wang, Zhang, Lu, Zhang, Yan, and Zhang]{wang-etal-2022-qadialmoe}
Longzheng Wang, Peng Zhang, Xiaoyu Lu, Lei Zhang, Chaoyang Yan, and Chuang Zhang.
\newblock {Q}a{D}ial{M}o{E}: Question-answering dialogue based fact verification with mixture of experts.
\newblock In Yoav Goldberg, Zornitsa Kozareva, and Yue Zhang, editors, \emph{Findings of the Association for Computational Linguistics: EMNLP 2022}, pages 3146--3159, Abu Dhabi, United Arab Emirates, December 2022. Association for Computational Linguistics.
\newblock \doi{10.18653/v1/2022.findings-emnlp.229}.
\newblock URL \url{https://aclanthology.org/2022.findings-emnlp.229}.

\bibitem[Wang et~al.(2023)Wang, Kordi, Mishra, Liu, Smith, Khashabi, and Hajishirzi]{wang2023selfinstruct}
Yizhong Wang, Yeganeh Kordi, Swaroop Mishra, Alisa Liu, Noah~A. Smith, Daniel Khashabi, and Hannaneh Hajishirzi.
\newblock Self-instruct: Aligning language models with self-generated instructions, 2023.

\bibitem[Wei et~al.(2024)Wei, Wang, Schuurmans, Bosma, Ichter, Xia, Chi, Le, and Zhou]{10.5555/3600270.3602070}
Jason Wei, Xuezhi Wang, Dale Schuurmans, Maarten Bosma, Brian Ichter, Fei Xia, Ed~H. Chi, Quoc~V. Le, and Denny Zhou.
\newblock Chain-of-thought prompting elicits reasoning in large language models.
\newblock In \emph{Proceedings of the 36th International Conference on Neural Information Processing Systems}, NIPS '22, Red Hook, NY, USA, 2024. Curran Associates Inc.
\newblock ISBN 9781713871088.

\bibitem[Wu et~al.(2023)Wu, Hu, and Feng]{wu2023enhancing}
Zirui Wu, Nan Hu, and Yansong Feng.
\newblock Enhancing structured evidence extraction for fact verification.
\newblock In \emph{The 2023 Conference on Empirical Methods in Natural Language Processing}, 2023.
\newblock URL \url{https://openreview.net/forum?id=b7ZJcAkjC3}.

\bibitem[Xu et~al.(2024)Xu, Sharaf, Chen, Tan, Shen, Durme, Murray, and Kim]{xu2024contrastive}
Haoran Xu, Amr Sharaf, Yunmo Chen, Weiting Tan, Lingfeng Shen, Benjamin~Van Durme, Kenton Murray, and Young~Jin Kim.
\newblock Contrastive preference optimization: Pushing the boundaries of llm performance in machine translation, 2024.

\bibitem[Yang et~al.(2018)Yang, Qi, Zhang, Bengio, Cohen, Salakhutdinov, and Manning]{yang-etal-2018-hotpotqa}
Zhilin Yang, Peng Qi, Saizheng Zhang, Yoshua Bengio, William Cohen, Ruslan Salakhutdinov, and Christopher~D. Manning.
\newblock {H}otpot{QA}: A dataset for diverse, explainable multi-hop question answering.
\newblock In Ellen Riloff, David Chiang, Julia Hockenmaier, and Jun{'}ichi Tsujii, editors, \emph{Proceedings of the 2018 Conference on Empirical Methods in Natural Language Processing}, pages 2369--2380, Brussels, Belgium, October-November 2018. Association for Computational Linguistics.
\newblock \doi{10.18653/v1/D18-1259}.
\newblock URL \url{https://aclanthology.org/D18-1259}.

\bibitem[Yao et~al.(2022)Yao, Zhao, Yu, Du, Shafran, Narasimhan, and Cao]{yao2022react}
Shunyu Yao, Jeffrey Zhao, Dian Yu, Nan Du, Izhak Shafran, Karthik Narasimhan, and Yuan Cao.
\newblock React: Synergizing reasoning and acting in language models.
\newblock \emph{arXiv preprint arXiv:2210.03629}, 2022.

\bibitem[Zhang et~al.(2024)Zhang, Zhang, and Zhou]{Zhang_Zhang_Zhou_2024}
Congzhi Zhang, Linhai Zhang, and Deyu Zhou.
\newblock Causal walk: Debiasing multi-hop fact verification with front-door adjustment.
\newblock \emph{Proceedings of the AAAI Conference on Artificial Intelligence}, 38\penalty0 (17):\penalty0 19533--19541, Mar. 2024.
\newblock \doi{10.1609/aaai.v38i17.29925}.
\newblock URL \url{https://ojs.aaai.org/index.php/AAAI/article/view/29925}.

\bibitem[Zhang et~al.(2022)Zhang, Ji, Zhang, and Passonneau]{zhang-etal-2022-contrastive-data}
Rui Zhang, Yangfeng Ji, Yue Zhang, and Rebecca~J. Passonneau.
\newblock Contrastive data and learning for natural language processing.
\newblock In Miguel Ballesteros, Yulia Tsvetkov, and Cecilia~O. Alm, editors, \emph{Proceedings of the 2022 Conference of the North American Chapter of the Association for Computational Linguistics: Human Language Technologies: Tutorial Abstracts}, pages 39--47, Seattle, United States, July 2022. Association for Computational Linguistics.
\newblock \doi{10.18653/v1/2022.naacl-tutorials.6}.
\newblock URL \url{https://aclanthology.org/2022.naacl-tutorials.6}.

\bibitem[Zhang and Gao(2023)]{zhang-gao-2023-towards}
Xuan Zhang and Wei Gao.
\newblock Towards {LLM}-based fact verification on news claims with a hierarchical step-by-step prompting method.
\newblock In Jong~C. Park, Yuki Arase, Baotian Hu, Wei Lu, Derry Wijaya, Ayu Purwarianti, and Adila~Alfa Krisnadhi, editors, \emph{Proceedings of the 13th International Joint Conference on Natural Language Processing and the 3rd Conference of the Asia-Pacific Chapter of the Association for Computational Linguistics (Volume 1: Long Papers)}, pages 996--1011, Nusa Dua, Bali, November 2023. Association for Computational Linguistics.
\newblock \doi{10.18653/v1/2023.ijcnlp-main.64}.
\newblock URL \url{https://aclanthology.org/2023.ijcnlp-main.64}.

\bibitem[Zhu et~al.(2023)Zhu, Si, Zhao, Zhu, Zhou, and He]{zhu-etal-2023-explain}
Yingjie Zhu, Jiasheng Si, Yibo Zhao, Haiyang Zhu, Deyu Zhou, and Yulan He.
\newblock {EXPLAIN}, {EDIT}, {GENERATE}: Rationale-sensitive counterfactual data augmentation for multi-hop fact verification.
\newblock In Houda Bouamor, Juan Pino, and Kalika Bali, editors, \emph{Proceedings of the 2023 Conference on Empirical Methods in Natural Language Processing}, pages 13377--13392, Singapore, December 2023. Association for Computational Linguistics.
\newblock \doi{10.18653/v1/2023.emnlp-main.826}.
\newblock URL \url{https://aclanthology.org/2023.emnlp-main.826}.

\end{thebibliography}

\appendix
\section{Design of Prompt Function}\label{design_prompt}

\begin{figure*}[t]
    \centering
    \includegraphics[width=1\linewidth]{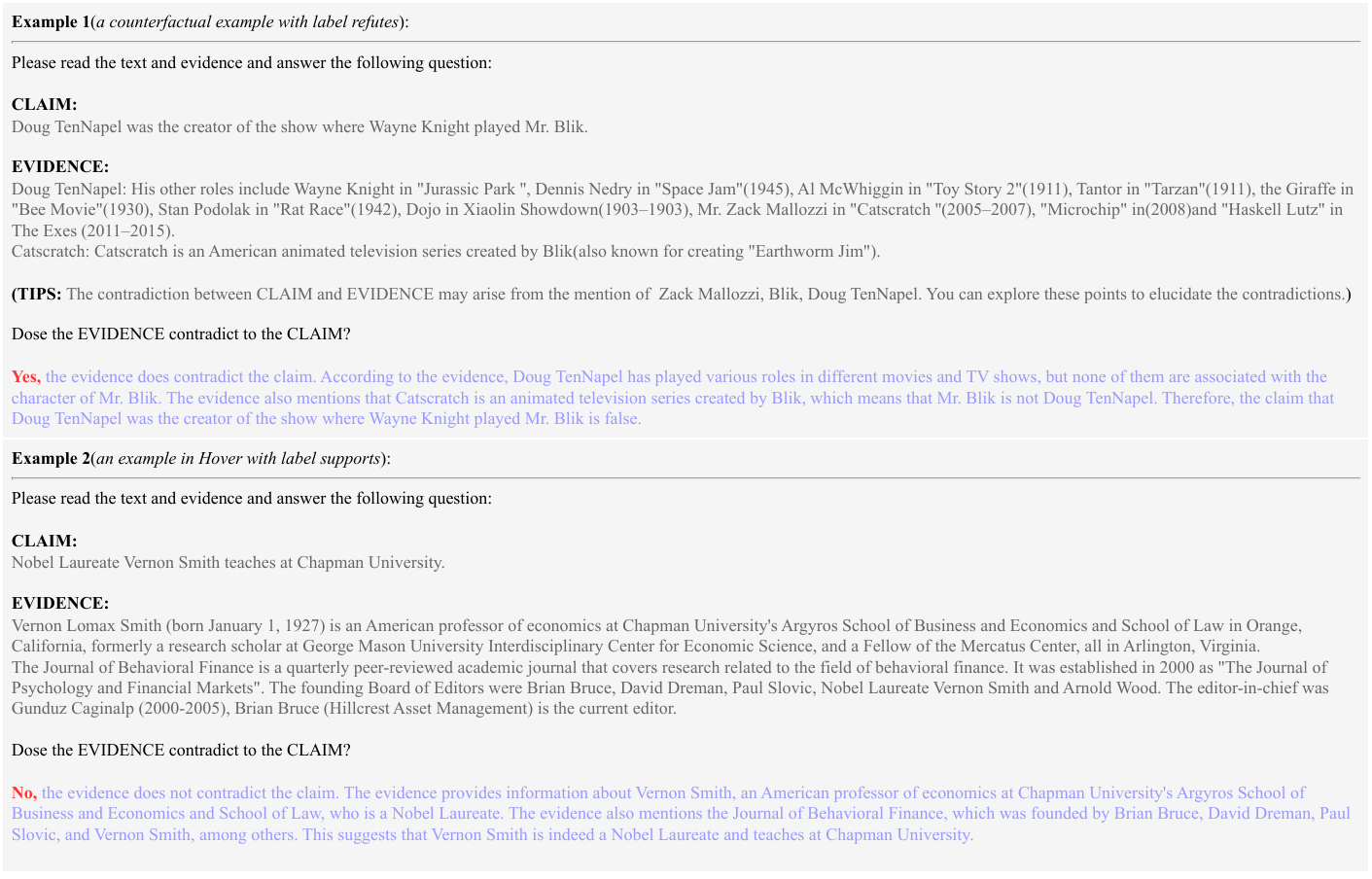}
    \caption[Short caption]{Two examples demonstrating how to generate prompts based on function $I$, with the first one coming from the counterfactual subset. Black font is used to combine information into coherent questions. Gray font represents $c$, $k$, and $tips$. Red font represents the label. Blue font indicates the explanation information generated by the model based on the question and label.}
    \label{prompt}
\end{figure*}

We demonstrate in Figure \ref{prompt} how to generate prompts using function $I$ given $c$, $k$, $tips$, and $y$. This method applies to both the data generation phase and the verification phase on the test set. During the verification phase, we only provide $c$ and $k$ and require the model to generate $y$ and $e$. In the data generation phase, if the data is from counterfactual subset with the label \blockquote{refutes} and includes editing information, in addition to providing $y$, we also provide $tips$ to assist the model in generating plausible explanations.


\section{Supplement of -exp}\label{sewp}
\label{sec:no_evi}

To gain a deeper understanding of the side effects of no evidence, we sampled from the test data and conducted case analyses. The specific results are shown in Table \ref{tab:no_evi}. In most cases, fine-tuning the model with labels only leads to a significant decrease in its generation capability, resulting in the generation of repetitive tokens(as shown in the first row of Table \ref{tab:no_evi}). In some cases, the model fails to comprehend contextual information or disregards certain contextual cues, and the model hallucinates, leading to the generation of fluent but incorrect information(as shown in the second and third rows of Table \ref{tab:no_evi}). In a few instances, the model even generates contradictory content, initially providing a negative conclusion but subsequently presenting a positive one(as shown in the fourth row of Figure \ref{tab:no_evi}).

\section{Statics of Dataset}
\label{section:sod}
We provide detailed statistical information for the FEVEROUS and HOVER, as shown in Table \ref{dataset_static}.

\section{Improved DPO Strategy}
\label{sec:poid}





\begin{figure*}[t]
\centering
\scriptsize
\begin{subfigure}{0.31\textwidth}
\includegraphics[width=\linewidth]{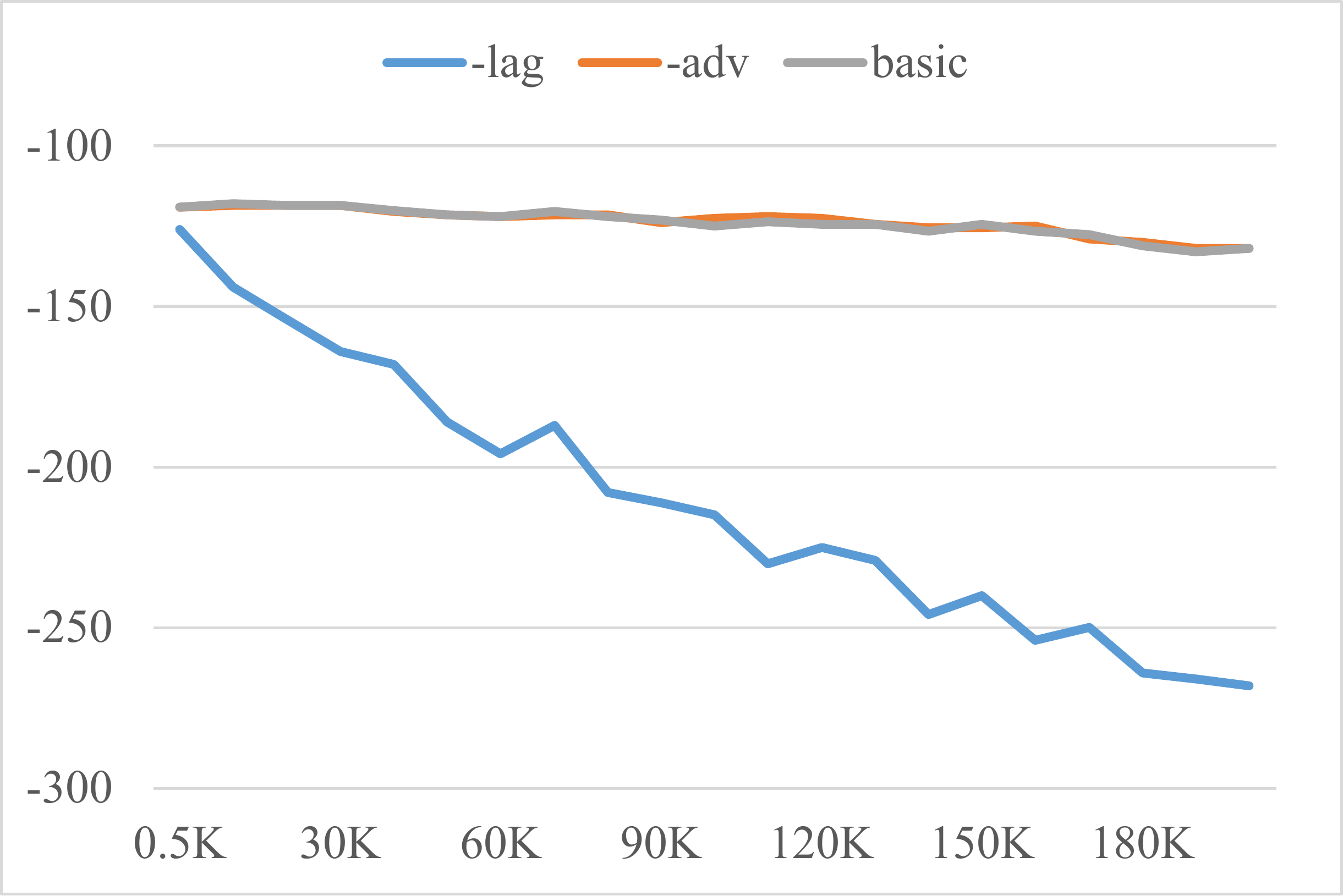}
\caption{\footnotesize Anthropic:\space Chosen}
\label{fig:hh_chosen}
\end{subfigure}
\begin{subfigure}{0.31\textwidth}
\includegraphics[width=\textwidth]{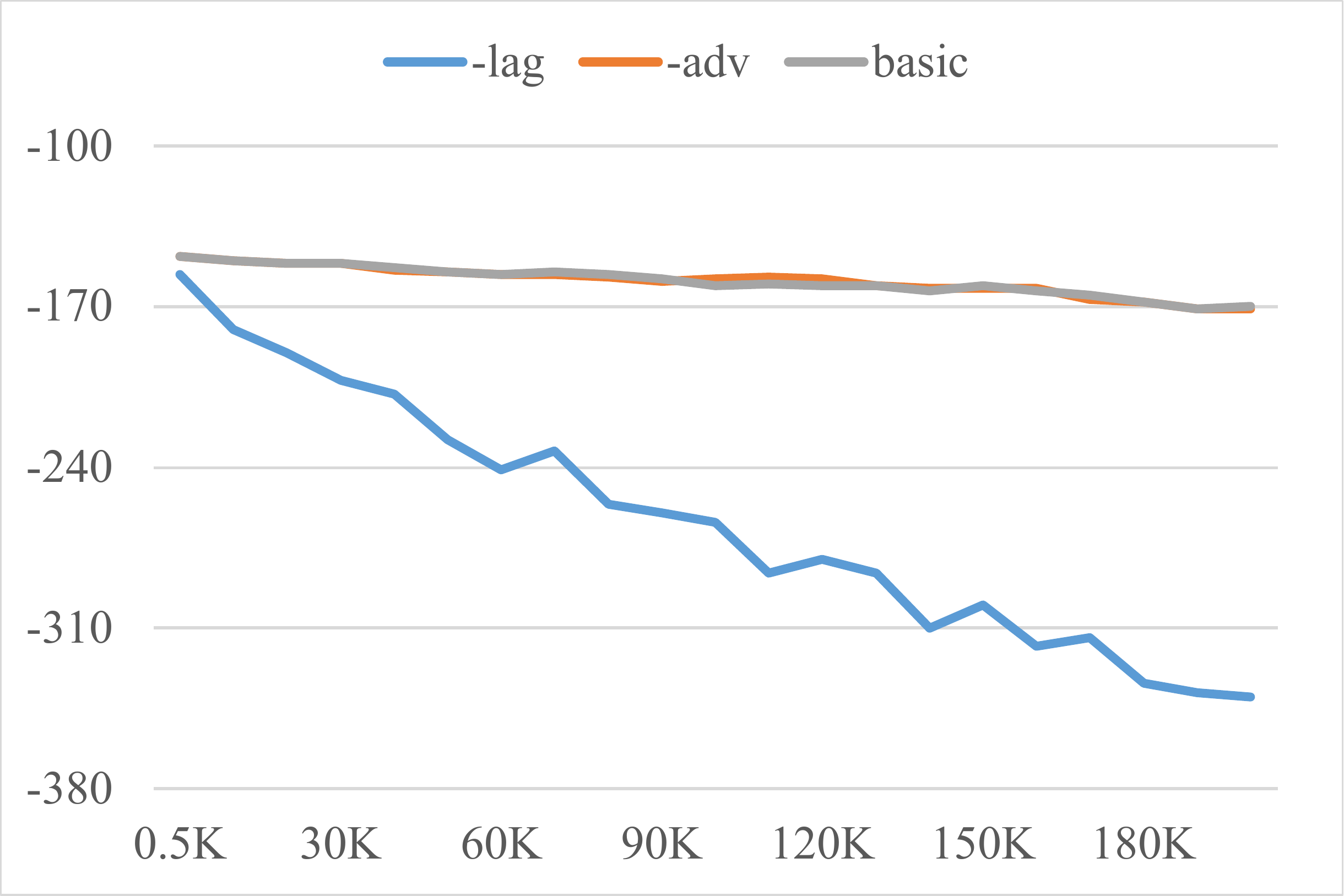}
\caption{\footnotesize Anthropic:\space Rejected}
\label{}
\end{subfigure}
\begin{subfigure}{0.31\textwidth}
\includegraphics[width=\linewidth]{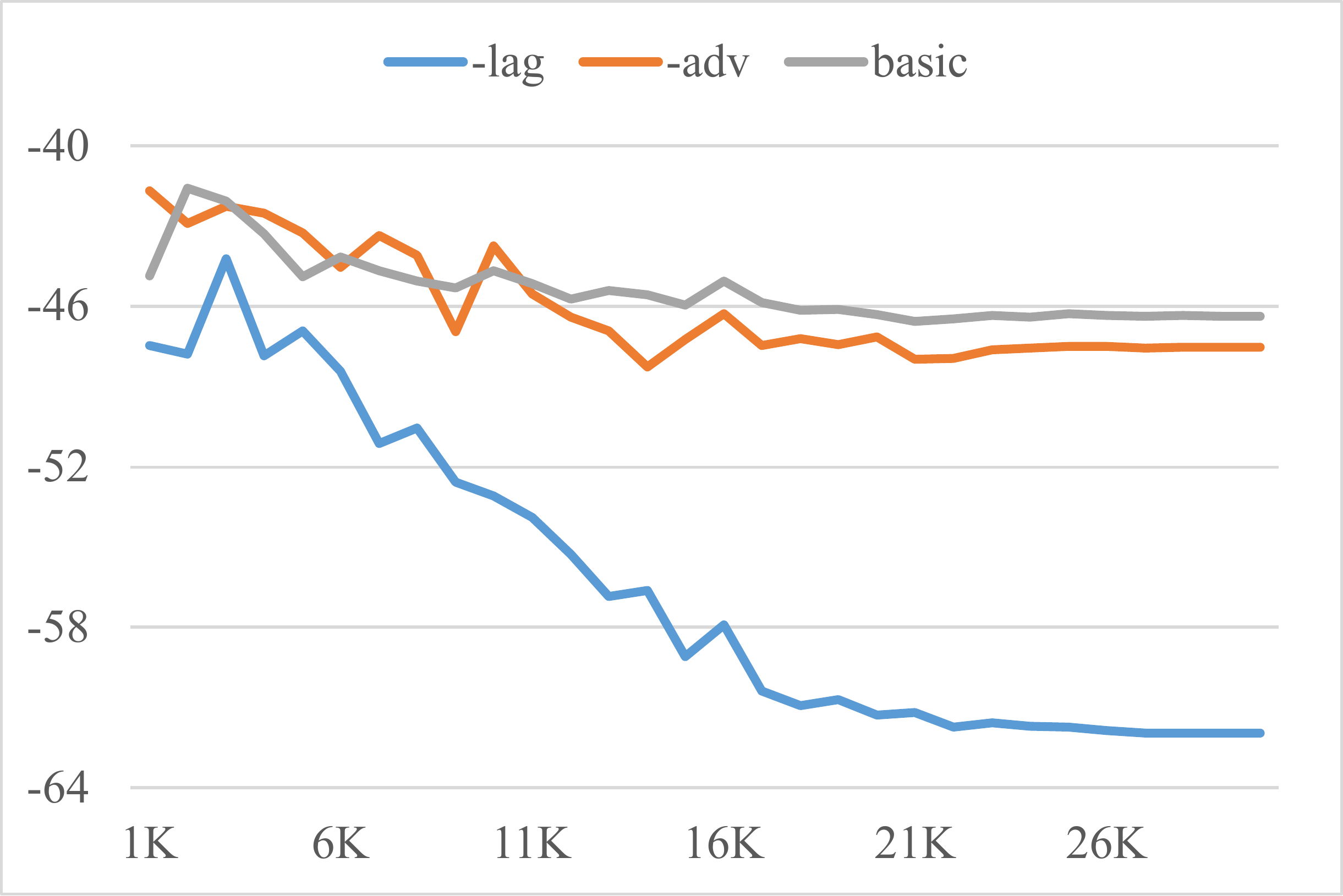}
\caption{\footnotesize FEVEROUS:\space Chosen}
\label{fig:feverous_chosen}
\end{subfigure}
\begin{subfigure}{0.31\textwidth}
\includegraphics[width=\textwidth]{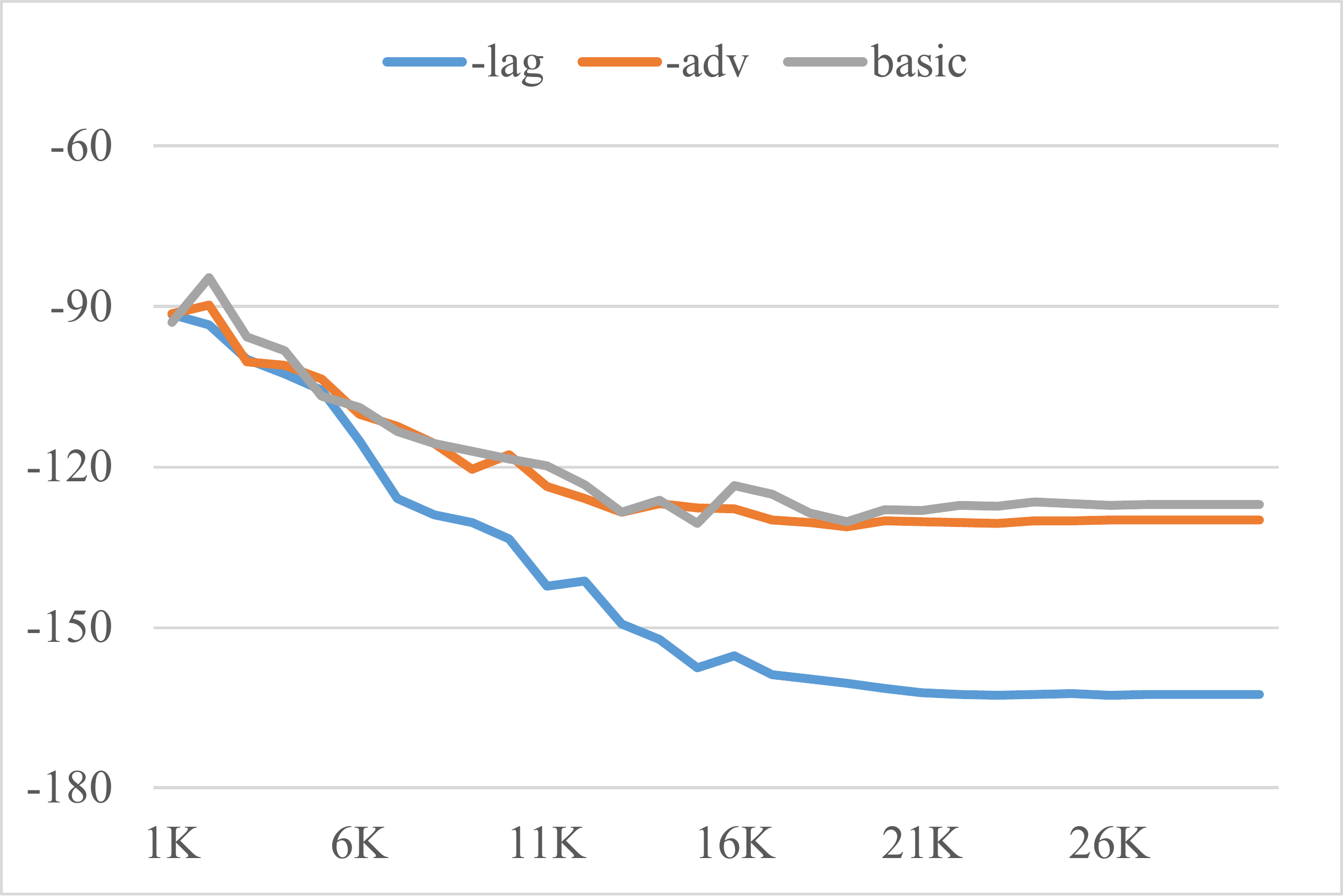}
\caption{\footnotesize FEVEROUS:\space Rejected}
\label{fig:feverous_rejected}
\end{subfigure}
\begin{subfigure}{0.31\textwidth}
\includegraphics[width=\linewidth]{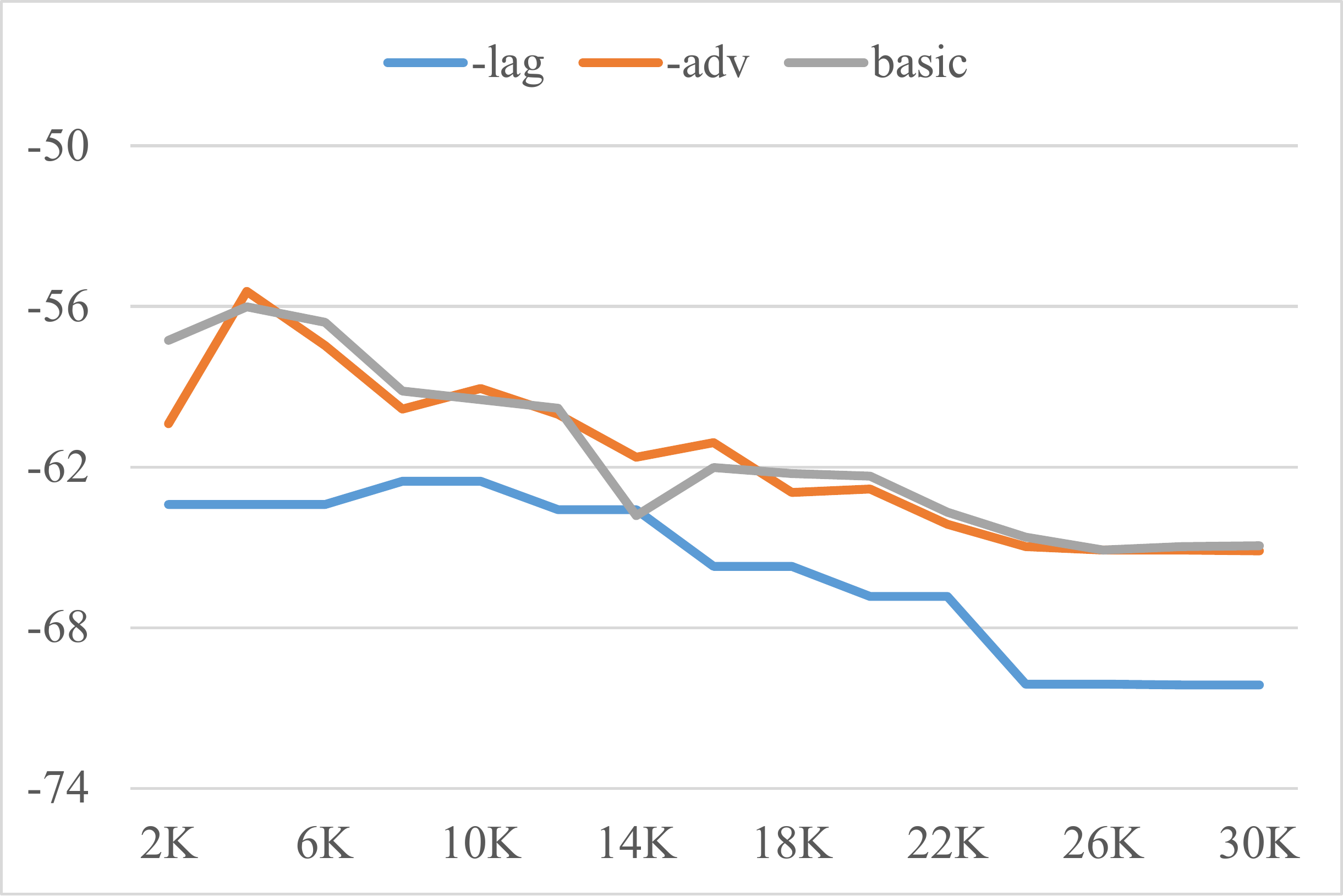}
\caption{\footnotesize HOVER:\space Chosen}
\label{fig:hover_chosen}
\end{subfigure}
\begin{subfigure}{0.31\textwidth}
\includegraphics[width=\textwidth]{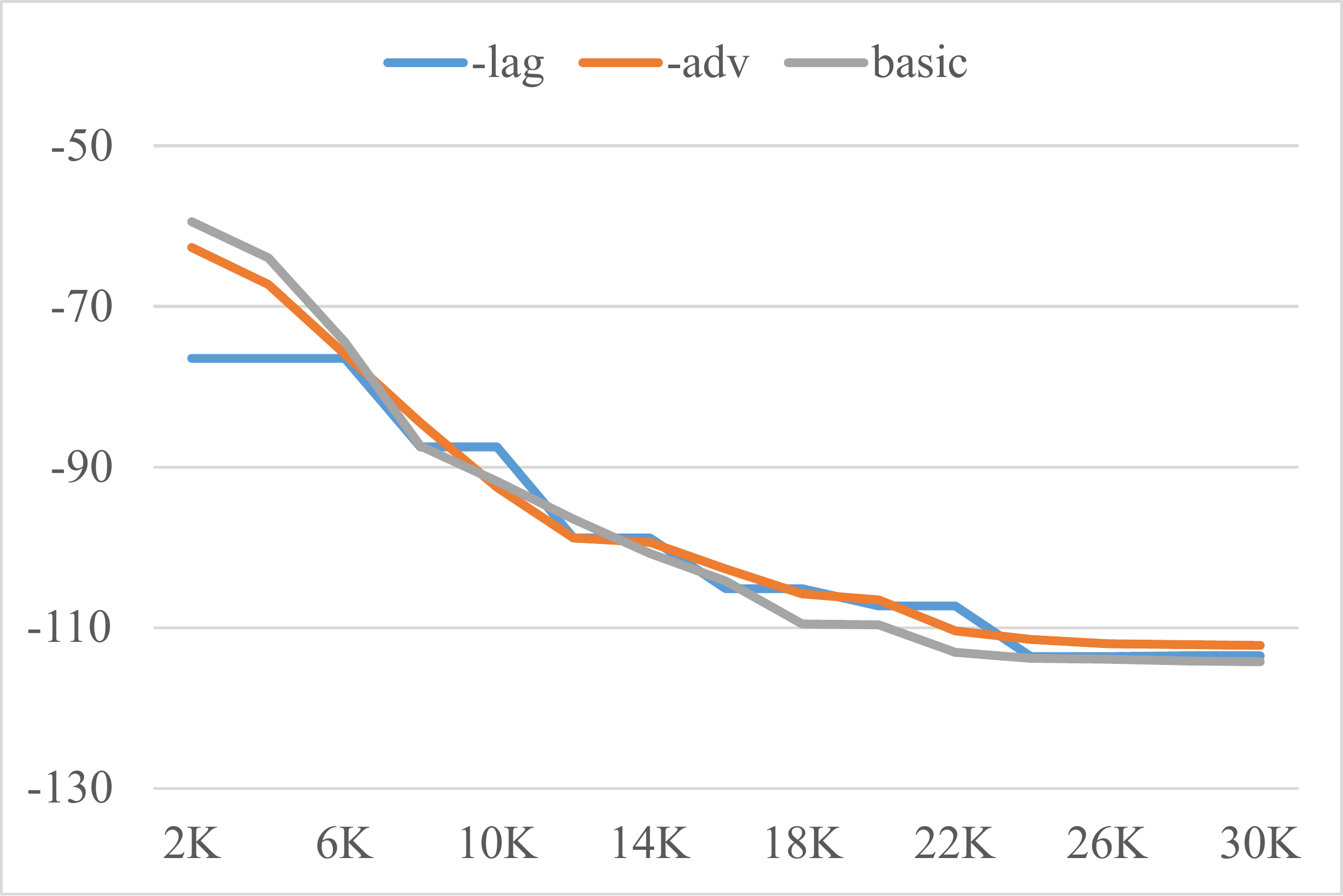}
\caption{\footnotesize HOVER:\space Rejected}
\label{fig:hover_rejected}
\end{subfigure}

\caption{The variation of text generation probabilities during the training process on the Anthropic, FEVEROUS, and HOVER validation datasets. The horizontal axis represents the training step, while the vertical axis represents the logarithm of the generation probability(logp/s). The terms "Chosen" and "Rejected" correspond to the positive and negative examples of self-instruction, respectively. We reported the results of three selected settings(-lag,-adv,baic) from the ablation experiments.}
\label{fig:poid}
\end{figure*}

During the training process, we observed a decrease in the generation probabilities of both $y_{chosen}$ and $y_{rejected}$, as shown in Figure \ref{fig:hh_chosen} and Figure \ref{fig:hh_chosen} (-lag).


This is inconsistent with our expectations, as we anticipate an increase in $\mathcal P(y_{chosen})$ after DPO fine-tuning. We speculate that the reason for this discrepancy is the lack of constraint on both $\mathcal P(y_{chosen})$ and $\mathcal P(y_{rejected})$. And even though, if the decrease rate of $\mathcal P(y_{rejected})$ relative to $\mathcal P(y_{chosen})$ is fast enough, the loss function will effectively decrease, which is precisely what we have observed from the experimental data. So we conduct improvement on DPO, as described in Section \ref{dpo_fine-tuning}.



We also evaluated the improved DPO on Anthropic, FEVEROUS and HOVER, as well as conducted ablation experiments -adv, to examine whether it can alleviate the problem of simultaneous decline in $y_{chosen}$ and $y_{rejected}$. As shown in Figure {fig:poid}, compared to DPO, the improved DPO alleviated the problem. This reflected in that as the training steps increase, the decline in $y_{chosen}$ is significantly reduced, while $y_{rejected}$ also maintains a downward trend.

\section{Parameter Setting}

We reported the parameter statistics of the fine-tuned model as shown in Table \ref{tab:parameter}. We employed the efficient fine-tuning approach for LLMs proposed by \cite{hu2022lora}, where we set the trainable parameter total(rank of trainable matrices) parameter $r$ and the Lipschitz constant parameter $alpha$ both to 128. It can be observed that the fine-tuning method we utilized only accounted for 0.99\% of the total parameters, resulting in an order of magnitude reduction compared to the fully fine-tuned model.

\begin{table}[t]
\scriptsize

\begin{subtable}{1\linewidth}
\begin{tabularx}{\linewidth}{
>{\hsize=0.1\hsize\linewidth=\hsize}X
>{\hsize=0.4\hsize\linewidth=\hsize}X 
>{\hsize=0.3\hsize\linewidth=\hsize}X 
>{\hsize=0.1\hsize\linewidth=\hsize}X 
}
\toprule
\tb{Split}             &\tb{Challenge}                 &\tb{Number}                    &\tb{Length}\\ \midrule
\multirow{7}{*}{train} & Combining Tables Text         & 65(10/55)                    & 162.33\\
                       & Entity Disambiguation         & 825(506/319)                 & 207.56\\
                       & Multi-hop Reasoning           & 4536(4027/509)               & 249.95\\
                       & Numerical Reasoning           & 643(126/517)                 & 170.78\\ 
                       & Search terms not in claim     & 444(257/187)                 & 192.71\\
                       & Other                         & 20415(10386/10029)           & 170.10\\ \cmidrule{1-4}
                       &\tb{total}                     &\tb{26928(15312/11616)}       &\tb{185.07}\\ \cmidrule{1-4}
\multirow{7}{*}{test}  & Combining Tables Text         & 7(3/4)                       & 194.42\\
                       & Entity Disambiguation         & 112(53/59)                   & 202.91\\
                       & Multi-hop Reasoning           & 459(376/83)                  & 245.14\\
                       & Numerical Reasoning           & 103(22/81)                   & 164.74\\ 
                       & Search terms not in claim     & 46(27/19)                    & 186.30\\
                       & Other                         & 2230(910/1305)               & 166.62\\ \cmidrule{1-4}
                       &\tb{total}                     &\tb{2962(1411/1551)}          &\tb{180.47}\\ \midrule
\end{tabularx}
\subcaption{FEVEROUS}
\label{tab:static_FEVEROUS}
\end{subtable}
\begin{subtable}{1\linewidth}
\begin{tabularx}{\linewidth}{
>{\hsize=0.1\hsize\linewidth=\hsize}X
>{\hsize=0.4\hsize\linewidth=\hsize}X 
>{\hsize=0.3\hsize\linewidth=\hsize}X 
>{\hsize=0.1\hsize\linewidth=\hsize}X 
}
\toprule
\tb{Split}             &\tb{Challenge}                 &\tb{Number}                    &\tb{Length}\\ \midrule
\multirow{7}{*}{train} & counterfactual                & 9738(3635/6103)              & 270.53\\
                       & 2\_hops                       & 9052(6496/2556)              & 304.32\\
                       & 3\_hops                       & 6084(3271/2813)              & 426.01\\
                       & 4\_hops                       &(1256/1779)                   & 554.12\\  \cmidrule{1-4}
                       &\tb{total}                     &\tb{27909(14658/13251)}       &\tb{346.22}\\  \cmidrule{1-4}
\multirow{4}{*}{test}  & 2\_hops                       & 1126(521/605)                & 302.89\\
                       & 3\_hops                       & 1835(968/867)                & 436.45\\
                       & 4\_hops                       & 1039(511/528)                & 563.34\\  \cmidrule{1-4}
                       &\tb{total}                     &\tb{4000(2000/2000)}          &\tb{431.82}\\  \midrule
\end{tabularx}
\subcaption{HOVER}
\label{tab:static_HOVER}
\end{subtable}
\caption{Detailed staticts of FEVEROUS and HOVER. The Number are presented as \blockquote{total count(number of positive samples/number of negative samples)}. Length represents the average token length.}
\label{dataset_static}
\end{table}

\begin{table}[t]
\scriptsize
\begin{tabularx}{\linewidth}{X X X X}
\toprule
\tb{model}  & \tb{total} &\tb{embeds}  &\tb{trainable} \\ \midrule
Bert-FC     & 1.09e+08   & 2.38e+07        & 1.09e+08\\ \midrule
LisT5       & 2.23e+08   & 2.47e+07        & 2.23e+08\\ \midrule
Ro-NLI      & 3.55e+08   & 5.20e+07        & 3.55e+08\\ \midrule
De-NLI      & 3.55e+08   & 5.20e+07        & 3.55e+08\\ \midrule
Ours        & 6.74e+09   & 1.31e+08        & 6.71e+07\\ \bottomrule
\end{tabularx}
\caption{Model parameter distribution. Total represents the total number of parameters. Embeds represents the number of parameters in the embedding layer. Trainable represents the number of trainable parameters.}
\label{tab:parameter}
\end{table}

{\fontsize{7}{7}\selectfont
\onecolumn
\newcommand{\tip}[1]{\textbf{\textit{\textcolor{gray}{#1}}}}
\renewcommand{\arraystretch}{1.5}
\begin{xltabular}{\textwidth}[t]{
>{\hsize=0.1\hsize\linewidth=\hsize}X
>{\hsize=0.9\hsize\linewidth=\hsize}X
}
\toprule[1pt]
\tb{Example 1}  & Label: \tb{Refutes}\\ \midrule
\tb{CLAIM} & YWHAG belongs to the 14-3-3 protein family found in only mammals(in heart muscles).\\
\tb{EVIDENCE}   & [14-3-3 protein] There are seven genes that encode seven distinct 14-3-3 proteins in most mammals(See Human genes below) and 13-15 genes in many higher plants, though typically in fungi they are present only in pairs. \\
\tb{Our Method} & Yes, the evidence provided appears to contradict the claim. According to the evidence, 14-3-3 proteins are found in both mammals and plants, specifically in the heart muscles of mammals. This suggests that YWHAG, which is a 14-3-3 protein, is not exclusive to mammals as claimed in the prompt.\\
\tb{-exp}       & Yes Yes Yes Yes Yes Yes Yes Yes Yes Yes Yes Yes Yes Yes Yes...\\ 

\toprule[1pt]
\tb{Example 2}  & Label: \tb{Supports}\\ \midrule
\tb{CLAIM} & Bill Persky(a TV director) was born to a Jewish family in 1931 and got married three times, where Joanna Patton was his third.\\
\tb{EVIDENCE}   & [Bill Persky] Bill Persky(born September 9, 1931) is an American director, screenwriter, producer, and actor for television. \newline
[Bill Persky] Persky was born to a American\_Jews family, the son of an estate auctioneer.\newline
[Bill Persky] Persky has been married three times.\newline
[Bill Persky] His third wife is New York advertising executive, Joanna Patton.\\
\tb{Our Method} & No, the evidence does not contradict the claim. The evidence provides information about Bill Persky's personal life, including his birthdate, occupation, and marital history. The claim states that Bill Persky was born to a Jewish family in 1931 and got married three times, with Joanna Patton being his third wife. The evidence supports this claim by providing the additional details that Persky was born to a Jewish family and has been married three times, including his marriage to Joanna Patton.\\
\tb{-exp}       & No No No No No, No evidence provided, it's a joke! This is an April Fool's Day joke! Bill Persky is a real person, and he is indeed a TV director, born to a Jewish family in 1931. But the part about him getting married three times, including to a woman named Joanna Patton, is the punchline. It turns out, Joanna Patton is a fictional character, and not a real person. So, there you have it, no evidence provided, just a harmless prank on April Fool's Day!\\ 

\toprule[1pt]
\tb{Example 3}  & Label: \tb{Supports}\\ \midrule
\tb{CLAIM} & Nijo Michihira had two wives and three children during his short lifetime of 47 years.\\
\tb{EVIDENCE}   & [Nijo Michihira] He married a daughter of Nijo Morotada and a daughter of Saionji Kin'aki.\newline 
[Nijo Michihira] The latter and he had son Nijo Yoshimoto, a daughter who was later a consort of Emperor Go-Daigo, and a son who was adopted by Tominokouji family and became known as Tominokouji Michinao.\newline
Nijo Michihira] Nijo Michihira(1288 – February 27, 1335) was a Japanese kugyo of the late Kamakura period.\\
\tb{Our Method} & No, the evidence does not contradict the claim. In fact, it supports the claim that Nijo Michihira had two wives and three children during his lifetime.The evidence states that Nijo Michihira married two women: a daughter of Nijo Morotada and a daughter of Saionji Kin'aki. This confirms that he had two wives. Additionally, the evidence mentions three children: Nijo Yoshimoto, a daughter who became a consort of Emperor Go-Daigo, and a son who was adopted by the Tominokouji family and became known as Tominokouji Michinao. This confirms that he had three children.Therefore, the evidence supports the claim that Nijo Michihira had two wives and three children during his lifetime of 47 years.\\
\tb{-exp}       & Yes Yes Yes, Yes. The evidence you provided contradicts the claim. According to the evidence, Nijo Michihira had only one wife, the daughter of Nijo Morotada, and had three children with her: a son named Nijo Yoshimoto, a daughter who later became a consort of Emperor Go-Daigo, and a son who was adopted by the Tominokouji family and became known as Tominokouji Michinao. This means that he had only one wife and three children, not two wives and three children as claimed in the original question. Additionally, Nijo Michihira lived for 47 years, as claimed in the original question, which is consistent with the evidence provided.\\ 

\toprule[1pt]
\tb{Example 4}  & Label: \tb{Supports}\\ \midrule
\tb{CLAIM} & Narazo Shimazu(born in 1888) went to Tokyo for foreign motorcycle races.\\
\tb{EVIDENCE}   & [Narazo Shimazu] Narazo Shimazu(1888–1973) founded Shimazu Motor Research Institute, that built the NS Motorcycle.\newline 
[Narazo Shimazu] Shimazu was born in Japan in 1888.\newline
[Narazo Shimazu] He went to Tokyo for races of foreign motorcycles.\\
\tb{Our Method} & No, the evidence does not contradict the claim. The evidence provides additional information about Narazo Shimazu\'s life, including his birth in Japan in 1888 and his founding of the Shimazu Motor Research Institute, which built the NS Motorcycle. The final sentence of the evidence, "He went to Tokyo for races of foreign motorcycles," supports the claim that Shimazu went to Tokyo for foreign motorcycle races. So, the evidence does not contradict the claim, but rather provides more details about Shimazu\'s life and interests.\\
\tb{-exp}       & No, no, no, no! The evidence provided contradict the claim. You are providing additional information about Narazo Shimazu, his academic background and his passion for motorcycles. Thank you for the clarification!\\ 

\bottomrule[1pt]
\caption{Some example displaying the side effect of without evidence. Inference prompt is the input of the model. Here, we omit the following question posed by the model in the prompt: "Does the Evidence contradict the CLAIM?". Our method refers to the correct results generated by the method we proposed. Without Evidence refers to the incorrect results generated by the model during testing using the -exp setting for fine-tuning.}
\label{tab:no_evi}
\end{xltabular}
\twocolumn
}

\end{document}